\documentclass[10pt,twocolumn,letterpaper]{article}

\usepackage[pagenumbers]{cvpr} % To force page numbers, e.g. for an arXiv version

\usepackage{amsmath,amsfonts}
\usepackage{algorithm}
\usepackage{algpseudocode}
\usepackage{amsthm}
\usepackage{array}
\usepackage{textcomp}
\usepackage{stfloats}
\usepackage{url}
\usepackage{verbatim}
\usepackage{graphicx}
\usepackage{multirow}
\usepackage{makecell}
\usepackage{tikz}

\newcommand\blfootnote[1]{% 
	\begingroup 
	\renewcommand\thefootnote{}\footnote{#1}% 
	\addtocounter{footnote}{-1}% 
	\endgroup 
}

\newcommand*\circled[1]{\tikz[baseline=(char.base)]{
    \node[shape=circle,draw,inner sep=0.5pt] (char) {#1};}}

\definecolor{cvprblue}{rgb}{0.21,0.49,0.74}
\usepackage[pagebackref,breaklinks,colorlinks,allcolors=cvprblue]{hyperref}

\def\paperID{11844} % *** Enter the Paper ID here
\def\confName{CVPR}
\def\confYear{2025}

\title{Towards Effective and Sparse Adversarial Attack on Spiking Neural Networks via Breaking Invisible Surrogate Gradients}

\author{Li Lun$^1$ \qquad Kunyu Feng$^2$ \qquad Qinglong Ni$^3$ \qquad Ling Liang$^1$ \qquad Yuan Wang$^1$ \\
Ying Li$^{3*}$ \qquad Dunshan Yu$^1$ \qquad Xiaoxin Cui$^{1*}$\\
$^1$School of Integrated Circuits, Peking University, Beijing, China\\
$^2$School of Software and Microelectronics, Peking University, Beijing, China\\
$^3$Institute of Microelectronics, Chinese Academy of Sciences, Beijing, China\\
{\tt\small \textbraceleft{}lunli, lingliang, wangyuan, yuds, cuixx\textbraceright{}@pku.edu.cn,}\\ {\tt\small feng\_ky21@stu.pku.edu.cn, \textbraceleft{}niqinglong, liying1\textbraceright{}@ime.ac.cn}
}

\begin{document}
\maketitle
\begin{abstract}
Spiking neural networks (SNNs) have shown their competence in handling spatial-temporal event-based data with low energy consumption. Similar to conventional artificial neural networks (ANNs), SNNs are also vulnerable to gradient-based adversarial attacks, wherein gradients are calculated by spatial-temporal back-propagation (STBP) and surrogate gradients (SGs). However, the SGs may be invisible for an inference-only model as they do not influence the inference results, and current gradient-based attacks are ineffective for binary dynamic images captured by the dynamic vision sensor (DVS). While some approaches addressed the issue of invisible SGs through universal SGs, their SGs lack a correlation with the victim model, resulting in sub-optimal performance. Moreover, the imperceptibility of existing SNN-based binary attacks is still insufficient. In this paper, we introduce an innovative potential-dependent surrogate gradient (PDSG) method to establish a robust connection between the SG and the model, thereby enhancing the adaptability of adversarial attacks across various models with invisible SGs. Additionally, we propose the sparse dynamic attack (SDA) to effectively attack binary dynamic images. Utilizing a generation-reduction paradigm, SDA can fully optimize the sparsity of adversarial perturbations. Experimental results demonstrate that our PDSG and SDA outperform state-of-the-art SNN-based attacks across various models and datasets. Specifically, our PDSG achieves 100\% attack success rate on ImageNet, and our SDA obtains 82\% attack success rate by modifying only 0.24\% of the pixels on CIFAR10DVS. The code is available at \url{https://github.com/ryime/PDSG-SDA}.
\end{abstract}
\blfootnote{$^*$ Corresponding Authors.}    
\section{Introduction}
\label{sec:intro}

Brain-inspired spiking neural networks (SNNs), as the third generation of neural network models, have attracted extensive attention in machine intelligence and neuromorphic computing \cite{maass1997networks, roy2019towards}. In contrast to floating-point activation in conventional artificial neural networks (ANNs), neurons in SNNs utilize binary spike sequences for communication. The sparse and event-driven natures of spikes endow SNNs with the abilities of asynchronous processing and low energy consumption \cite{li2024brain, rathi2023exploring}. Nowadays, SNN-based neuromorphic hardware has emerged as low-power and high-performance edge computing devices in resource-constrained scenarios, such as TrueNorth \cite{akopyan2015truenorth}, Loihi \cite{davies2018loihi, orchard2021efficient}, Tianjic \cite{pei2019towards}, and Darwin \cite{ma2017darwin, ma2024darwin3}.
\begin{figure}[tb]
  \centering
  \includegraphics[height=5cm]{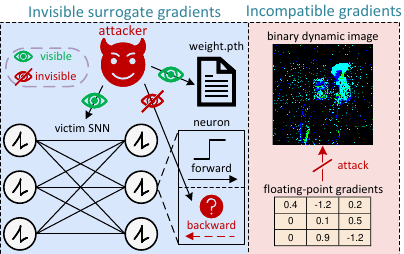}
  \caption{Illustration of the challenges of attacking SNNs. The invisible SGs hinder the attacker to perform gradient-based attacks. The incompatible gradients describe that the floating-point gradients are difficult to be converted to binary perturbations.}
  \label{fig:challenge}
\end{figure}

In practical applications, the security and robustness of the neural network are required to be concerned \cite{zhou2022adversarial, han2023interpreting}. Prior studies \cite{szegedy2013intriguing, goodfellow2014explaining, madry2018towards} have highlighted the vulnerability of ANNs to gradient-based adversarial attacks, which craft adversarial examples with imperceptible perturbations to fool the model. In SNNs, gradients can be calculated using spatial-temporal back-propagation (STBP) \cite{wu2018spatio}, which adopts the surrogate gradient (SG) to circumvent the non-differentiability of the Heaviside function \cite{neftci2019surrogate}. However, since the inference only requires the structure and trained parameters of the model, the attacker might not obtain the SG of an inference-only model. As the shape and hyper-parameters of the SG significantly influence the attack success rate \cite{xu2022securing}, choosing an appropriate SG is crucial for enhancing the effectiveness of the attack. Moreover, SNNs with inherent spatial-temporal characteristics excel at handling dynamic images, represented by events captured by the dynamic vision sensor (DVS) \cite{li2022neuromorphic}. Typically, these events are aggregated into binary frames to maintain compatibility with neuromorphic hardware \cite{zhang2024anp, yao2024spike, zhong2024paicore}, making gradient-based attacks inapplicable to binary dynamic images due to different input formats \cite{yao2024exploring}. These challenges are depicted in \cref{fig:challenge}.

To address the issue of invisible SGs, methods such as RGA \cite{bu2023rate} and HART \cite{hao2024threaten} adopted customized SGs. Nonetheless, their SGs are universal and lack a direct correlation with the victim model. Even when the attacker fortunately obtains and utilizes the SG used during the training phase \cite{sharmin2020inherent}, the performance of the attack remains unstable, as the training-phase SG is dedicated to optimizing the model's parameters rather than the gradient of the input. Consequently, establishing a connection between the attacking-phase SG and the trained model would significantly advance the adversarial attacks. Additionally, in the context of binary dynamic images, the adversarial perturbations should be sparse enough to evade detection, since the aggregated frames are spatially sparse \cite{wang2024HARDVS}. Although existing attack paradigms have successfully converted floating-point gradient to binary perturbations \cite{liang2021exploring, buchel2022adversarial, yao2024exploring}, they often suffer from gradient vanishing, and the imperceptibility of perturbations remains unsatisfactory. Therefore, an effective sparse attack method is required to thoroughly investigate the robustness of SNNs.

In this paper, we propose the potential-dependent surrogate gradient (PDSG) in adversarial attacks on SNNs. The PDSG can adapt to various models due to its shape dependent on the distribution of the membrane potential, thereby enhancing the threats of adversarial attacks across static and dynamic datasets. We also devise a paradigm for attacking binary dynamic images, named sparse dynamic attack (SDA). SDA iteratively generates significant and removes redundant perturbations, fully optimizing the imperceptibility of adversarial perturbations. The main contributions of our paper are summarized as follows:
\begin{itemize}
    \item We propose the potential-dependent surrogate gradient to achieve more representative gradients in adversarial attacks. The PDSG establishes a connection between the SG and the victim model through the run-time distribution of membrane potential. To the best of our knowledge, this is the first time that an adaptive SG is adopted in adversarial attacks on SNNs.
    \item We introduce the sparse dynamic attack on binary dynamic images, which combines the gradient and the finite difference to craft sparse yet powerful adversarial examples. Our SDA carefully selects the most desirable pixels to attack, effectively improving the sparsity of adversarial perturbations while maintaining high attack success rate.
    \item We conduct extensive experiments on various datasets and SNN models to substantiate the effectiveness of our PDSG and SDA. On ImageNet dataset, the PDSG achieves 100\% attack success rate without knowing the SG of the model. For binary dynamic images on CIFAR10DVS dataset, the SDA obtains 82\% attack success rate with only 0.24\% pixels perturbed.
\end{itemize}
\section{Related Works}
\label{sec:related}

\subsection{High Performance Spiking Neural Networks}
There are two main types of learning algorithms for achieving high-performance SNNs: converting ANNs to SNNs (ANN2SNN) 
\cite{li2021free, bu2022optimal, you2024spikeziptf} and directly training SNNs \cite{zhou2024direct, guo2023direct}. As ANN2SNN methods have been demonstrated to be more vulnerable to adversarial attacks \cite{sharmin2019comprehensive, sharmin2020inherent}, we only consider directly-trained SNNs in this paper. The widely used algorithm for directly training deep SNNs is STBP \cite{wu2018spatio}. STBP clarifies the propagation process across the spatial and temporal dimension, and the SG is adopted to mitigate the non-differentiable problem of the Heaviside function. Various shapes and parameters of SG are also investigated \cite{lian2023learnable, li2021differentiable, wang2023adaptive, zenke2021remarkable}. To effectively train deep SNNs, threshold-dependent batch normalization (tdBN) \cite{zheng2021going} is proposed to balance the distribution of the pre-activations and the neuronal threshold, thereby stabilizing the gradient flow. Implementing optimization algorithms, current directly-trained SNNs exhibit low inference latency and comparable accuracy to ANNs on both static and event-based datasets \cite{deng2022temporal, fang2021deep, yao2022glif, duan2022temporal, yao2023attention, yao2024spikedriven}.

\subsection{Adversarial Attacks and Defenses on Directly-trained SNNs}
Directly-trained SNNs are vulnerable to adversarial examples crafted by STBP. Sharmin {\it{et al.}} proposed a gradient propagation algorithm for Poisson encoding on static images and demonstrated the power of adversarial attack based on STBP \cite{sharmin2020inherent}. To optimize the gradient flow, RGA \cite{bu2023rate} and HART \cite{hao2024threaten} investigated the rate and temporal information of SNNs, and generated more potent adversarial examples through changing the calculation process of STBP. For binary dynamic inputs of SNNs, DVS-attacks \cite{marchisio2021dvs} first developed various searching methods on DVS data to fool the SNNs. To generate sparse perturbations, spike-compatible gradient (SCG) \cite{liang2021exploring} converted floating-point gradients to binary spike through probabilistic sampling. SpikeFool \cite{buchel2022adversarial} adapted the SparseFool \cite{modas2019sparsefool} from ANNs to SNNs by rounding the sparse floating-point perturbations to binary values. In addition, Gumbel-softmax attack (GSAttack) \cite{yao2024exploring} directly perturbed the raw event data through the Gumbel-softmax technique.

To mitigate the impact of adversarial attacks, several defense strategies inspired by ANN-based methods are proposed \cite{ozdenizci2024adversarially}. Certified training is extended to SNNs through investigating the interval bound propagation \cite{liang2022toward} and randomized smoothing \cite{mukhoty2024certified}. The weight regularization \cite{ding2022snn} and gradient regularization \cite{liu2024enhancing} are adopted in adversarial training on SNNs. To reach bio-plausible robustness, the dynamics of neurons is reformed through stochastic gating \cite{ding2024enhancing} and lateral inhibition \cite{zhang2023take}. The effects of noise filters for DVS inputs were also discussed \cite{marchisio2021dvs, marchisio2021r}. Moreover, various works investigated the inherent robustness of SNNs, such as leakage factor \cite{el2021securing, ding2024robust}, coding schemes \cite{kundu2021hire, li2022comparative}, and firing threshold \cite{el2021securing}.
\section{Preliminaries}
\label{sec:pre}

\subsection{Leaky-Integrate-and-Fire Neuron Model}

\begin{figure*}[tb]
  \centering
  \includegraphics[height=4.95cm]{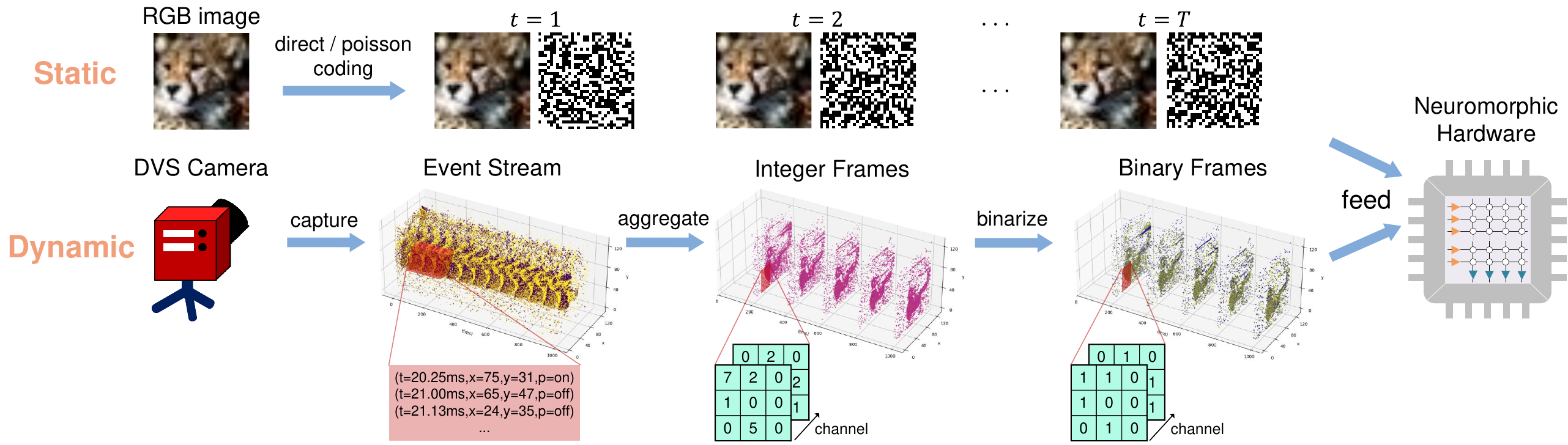}
  \caption{Illustration of the pre-processing procedure for both static and dynamic input of SNNs. For static inputs, the RGB image is encoded by direct or Poisson encoding to extend the temporal dimension. For dynamic inputs, event stream is captured by DVS, where $t$ denotes the time of the event, $x,y$ is the coordinates, and $p$ is the polarity. Then the event stream is aggregated into several integer frames, and each polarity corresponds to a channel. The integer frames will be further binarized to binary frames for hardware compatibility.}
  \label{fig:preprocess}
\end{figure*}

In this paper, we adopt the widely used leaky-integrated-and-fire (LIF) neuron model \cite{gerstner2014neuronal} in SNNs. Considering the iterative expression of the LIF model \cite{wu2019direct}, the membrane potential of the $l$-th layer and $t$-th timestep is updated by:
\begin{equation}
    \boldsymbol{u}^l[t] = \tau \boldsymbol{u}^l[t-1](1-\boldsymbol{s}^l[t-1]) + \boldsymbol{W}^l\boldsymbol{s}^{l-1}[t] + \boldsymbol{b}^l.
    \label{eq:lif}
\end{equation}

$\boldsymbol{u}$ represents the membrane potential of the neuron, $\tau$ denotes the leakage factor, and $\boldsymbol{W}^l$ and $\boldsymbol{b}^l$ are weight and bias. When the membrane potential reaches the threshold $V_{th}$, the neuron will fire a spike $\boldsymbol{s}^{l-1}$ through the Heaviside function and trigger the reset mechanism. Here we consider the hard reset mechanism, which directly resets the membrane potential to $0$. The firing function is described as:
\begin{equation}
    \boldsymbol{s}^l[t] = h(\boldsymbol{u}^l[t]-V_{th}) = \begin{cases}
        1, &\boldsymbol{u}^l[t] \ge V_{th} \\
        0, &\text{otherwise}
    \end{cases}.
    \label{eq:fire}
\end{equation}

The LIF model describes the spatial-temporal characteristic of the neuron. The firing and reset mechanisms introduce the non-linearity to the neuron, enabling SNNs to perform complex tasks.

\subsection{Spatial-Temporal Backpropagation}
Based on the neurodynamics of the LIF neuron model, STBP algorithm \cite{wu2018spatio} demonstrates that the gradients of the spikes contain both spatial and temporal terms. The gradient of the loss function $\mathcal{L}$ with respect to the spikes $\boldsymbol{s}^l[t]$ in the $l$-th layer is calculated by:
\begin{align}
    \frac{\partial \mathcal{L}}{\partial \boldsymbol{s}^l[t]} = &\frac{\partial \mathcal{L}}{\partial \boldsymbol{s}^{l+1}[t]}\frac{\partial \boldsymbol{s}^{l+1}[t]}{\partial \boldsymbol{u}^{l+1}[t]}\frac{\partial \boldsymbol{u}^{l+1}[t]}{\partial \boldsymbol{s}^l[t]} + \notag \\
    &\frac{\partial \mathcal{L}}{\partial \boldsymbol{s}^l[t+1]}\frac{\partial \boldsymbol{s}^l[t+1]}{\partial \boldsymbol{u}^l[t+1]}\frac{\partial \boldsymbol{u}^l[t+1]}{\partial \boldsymbol{s}^l[t]}.
    \label{eq:stbp}
\end{align}    

The SG is adopted to address the non-differentiable problem of the firing function in \cref{eq:fire}, where the derivative $\partial \boldsymbol{s}^{l}[t] / \partial \boldsymbol{u}^{l}[t]$ can not be directly calculated. The shape of the SG plays a crucial role in the representation of the gradient. A sharp SG can induce gradient vanishing, wherein most gradients become zero; conversely, a flat SG can cause gradient mismatch, indicating that the gradient fails to accurately reflect the trend of loss changes \cite{lian2023learnable, guo2022recdis}. This phenomenon implies that the effectiveness of gradient-based adversarial attacks is highly dependent on the SG.

\subsection{Input Pre-processing}
As SNNs contain calculations across the temporal dimension, the inputs of SNNs must be transformed to align with their spatial-temporal characteristics. For static images, common coding schemes include direct coding and Poisson coding \cite{kim2022rate}. In direct coding, the image at every timestep is identical to the input image. Poisson coding generates a binary image at each timestep, following a Bernoulli distribution where the probability is determined by the normalized pixel values. These coding schemes are denoted as:
\begin{align}
    x[t] = \begin{cases}
        x, &\text{direct coding}\\
        x > rand(0,1), &\text{Poisson coding}
    \end{cases}.
    \label{eq:coding}
\end{align}

For dynamic data (\eg~, in DVS-Gesture \cite{amir2017low} dataset), the event stream is required to be aggregated into integer frames, with the pixel value representing the count of events within the time interval \cite{yao2021temporal}. As some neuromorphic processors are capable of processing multi-bit inputs, like Tianjic \cite{pei2019towards} and Loihi2 \cite{orchard2021efficient}, they can handle both static images and integer dynamic images. However, most neuromorphic processors only accept binary spike inputs \cite{basu2022spiking, cho20192048, zhang2024anp, kuang202164k}, which aligns with the fundamental concept of SNNs. In these cases, binary frames for dynamic images are suitable for these processors. The pre-processing procedure is illustrated in \cref{fig:preprocess}. In this paper, we consider adversarial attacks on all these input scenarios for SNNs.

\subsection{Adversarial Attack}

The objective of an adversarial attack is to generate imperceptible perturbations which can fool the classifier. Given a SNN classifier $f$, we consider a benign image $\boldsymbol{x}$ with its corresponding label $y$. In this paper, we focus on the untargeted attack, where the attacker aims to change the classification result to any other label.  The optimization problem of the untargeted attack is denoted as:
\begin{equation}
    \text{argmax}_{\boldsymbol{\delta}} \mathcal{L}(f(\boldsymbol{x}+\boldsymbol{\delta}), y),\quad \text{subject to } ||\boldsymbol{\delta}||_p \le \epsilon.
    \label{adversarial_attack}
\end{equation}

Here $\boldsymbol{\delta}$ is the perturbation, and $||\boldsymbol{\delta}||_p$ is the $\ell_p$-norm of the perturbation. We denote $\boldsymbol{x}_{adv}=\boldsymbol{x}+\boldsymbol{\delta}$ as the adversarial example. For static images and integer dynamic frames, consistent with common adversarial attacks on ANNs, we adopt the $\ell_{\infty}$-norm to limit the maximum absolute value of perturbations. For binary dynamic frames, we use the $\ell_{0}$-norm to limit the number of modified pixels, which also represents the sparsity of adversarial perturbations.

In adversarial attacks on ANNs, there are two simple yet effective $\ell_{\infty}$-norm attack algorithms: fast gradient sign method (FGSM) \cite{goodfellow2014explaining} and projected gradient descent (PGD) \cite{madry2018towards}. FGSM leverages the sign of the gradient of the input to craft adversarial examples, which is denoted as:
\begin{equation}
    \boldsymbol{x}_{adv} = \boldsymbol{x} + \epsilon \cdot \text{sign}(\frac{\partial \mathcal{L}(f(\boldsymbol{x}), y)}{\partial \boldsymbol{x}}).
    \label{eq:fgsm}
\end{equation}

PGD iteratively performs the FGSM with a small step size $\alpha$. In the $k$-th iteration, the input is projected onto the space of the $\epsilon$-$\ell_\infty$ neighborhood by $\prod_{\epsilon}$, denoted as:
\begin{equation}
    \boldsymbol{x}^{k+1} = \prod_{\epsilon} \{\boldsymbol{x}^k + \alpha \cdot \text{sign}(\frac{\partial \mathcal{L}(f(\boldsymbol{x}^k), y)}{\partial \boldsymbol{x}^k})\}.
    \label{eq:pgd}
\end{equation}
\section{Methods}
\label{sec:methods}
\begin{figure*}[tb]
  \centering
  \includegraphics[height=4.8cm]{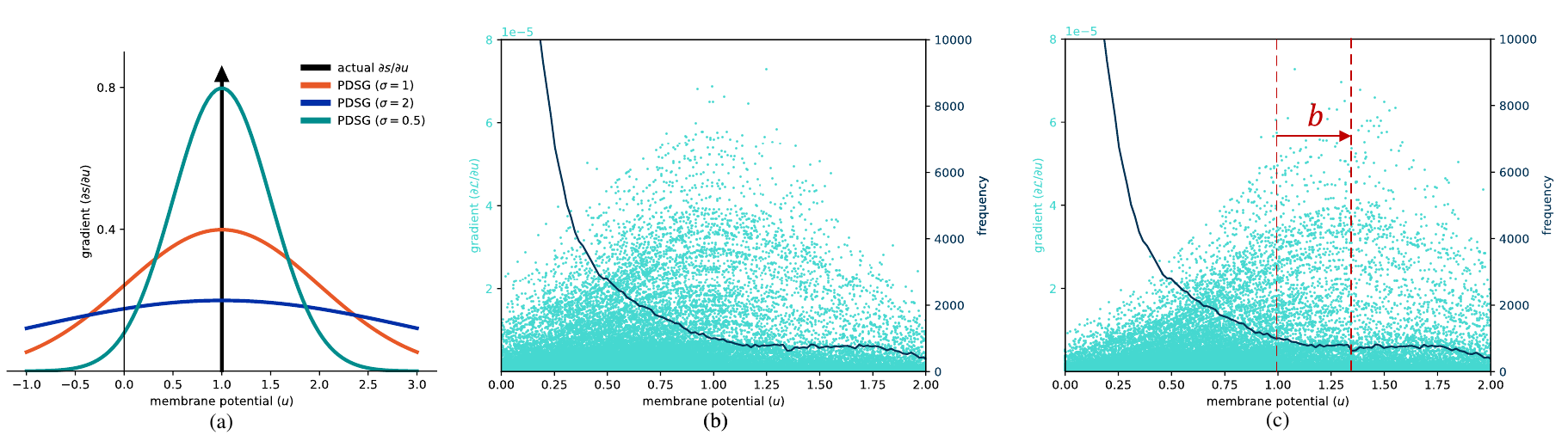}
  \caption{(a) Illustration of our PDSG under different distributions of membrane potential. (b) The scatter diagram of the gradients and the frequency curve of membrane potential in the penultimate layer of ResNet18. The gradients cluster on the left side of the threshold, causing imbalanced gradients. (c) The scatter diagram of the gradients after calibration. The distribution of the gradients is balanced around the threshold, urging the attack to pay equal attention to the gradients on the both sides of the threshold.}
  \label{fig:pdsg}
\end{figure*}

To address the challenge of invisible SGs when attacking SNNs, we first introduce the potential-dependent surrogate gradient ({\bf{PDSG}}). Our PDSG incorporates the variance of membrane potential into the shape of SG, thereby adapting to diverse distribution of models. Furthermore, we describe the sparse dynamic attack ({\bf{SDA}}) to generate sparse perturbations in attacking binary dynamic images. The generation-reduction paradigm of our SDA effectively improves the imperceptibility of adversarial perturbations.

\subsection{Potential-Dependent Surrogate Gradient}
\label{sec:pdsg}
Due to the non-differential problem of the firing function in SNNs, the attacker is difficult to perform the attack without knowledge of the SG. Selecting an SG blindly or randomly cannot guarantee the performance of the attack. We first explore the relationship between the distribution of membrane potential and the SG to establish our PDSG. Subsequently, we identify a distribution shift in the PDSG and continue to calibrate our PDSG through right-shifting.

{\bf{Derivation of PDSG.}} The zeroth-order method is instrumental in estimating the gradient within a specific distribution \cite{liu2020primer, mukhoty2023direct}. Therefore, we adopt the two-point zeroth-order method to the firing function as the foundation of our PDSG, which is expressed as:
\begin{equation}
    G^2(u;z,\delta)=\frac{h(u+z\delta-V_{th}) - h(u-z\delta-V_{th})}{2\delta}z.
    \label{eq:zeroth_order}
\end{equation}

Here, $z$ is sampled from a distribution $\lambda$, and $\delta$ is a constant smooth parameter. Then, the surrogate gradient can be calculated by the expectation of the two-point zeroth-order:
\begin{equation}
    \frac{\partial s}{\partial u} = \mathbb{E}_{z \sim \lambda}[G^2(u;z,\delta)]=\int_{\frac{|u-V_{th}|}{\delta}}^{\infty} \frac{|z|}{\delta}\lambda(z)dz.
    \label{eq:surrogate}
\end{equation}

Consequently, we begin by investigating the distribution $\lambda$ and the selection of $\delta$. Since $\Delta u = z\delta$ can be viewed as a change to the origin membrane potential $u$, we treat $u'=u+z\delta$ as the perturbed potential. According to \cref{eq:lif}, the value of potential depends solely on the spikes from the previous layer when the weight and bias are fixed. The firing rate should remain stable after perturbed, as a severe change of the firing rate can lead to attacks being detected in terms of power consumption \cite{krithivasan2022efficiency}. Therefore, $u'$ and $u$ can be approximately considered to be in the same distribution. Assuming that the membrane potential follows a normal distribution $\mathcal{N}(\mu, \sigma^2)$, then $z \sim \mathcal{N}(\frac{\mu-u}{\delta}, \frac{\sigma^2}{\delta^2})$. Given that the two-point zeroth-order estimation requires $\mathbb{E}(z)=0$ and $\mathbb{E}(z^2)=1$, we set $\delta=\sigma$, ensuring that $z \sim \mathcal{N}(0, 1)$ holds approximately when $u\approx\mu$. Substituting $\lambda(z)$ into \cref{eq:surrogate}, our PDSG is formulated as (a detailed derivation is provided in the Appendix):
\begin{equation}
    \frac{\partial \boldsymbol{s}^{l}[t]}{\partial \boldsymbol{u}^{l}[t]} = \frac{1}{\sqrt{2\pi}\boldsymbol{\sigma}}\text{exp}(-\frac{(\boldsymbol{u}^{l}[t]-V_{th})^2}{2\boldsymbol{\sigma}^2}).
    \label{eq:pdsg}
\end{equation}

We depict the PDSG under diverse distributions of membrane potential in \cref{fig:pdsg}(a). A large deviation implies that the membrane potential is more dispersed, and the possible changes are larger, necessitating a flatter surrogate gradient to encompass a wider range of membrane potential, and vice versa. Consequently, our PDSG can adapt to various distributions of membrane potential while disregarding the training-phase SG of the models.

{\bf{Calibration.}} \cref{eq:pdsg} approximately holds when $u\approx\mu$. However, our interest lies in the surrogate function near the threshold where $u$ significantly deviates from $\mu$, as the firing rate of neurons is far less than 50\% \cite{fang2021deep, wang2023adaptive}. We depict the distribution of the gradients in the penultimate layer of ResNet18 in \cref{fig:pdsg}(b). The distribution of membrane potential is not symmetric about $V_{th}$, inducing dominant gradients clustering on the left side of the threshold. In this case, the attack will disproportionately focus the gradients at $u < V_{th}$, potentially causing an increasing firing rate and neglecting gradients on the other side of the threshold. Since an increasing firing rate can lead to saturated gradients during attacks \cite{liang2021exploring}, the gradients around the threshold are required to be balanced.

To address this issue, we calibrate our PDSG through right-shifting the SG by bias $b$ in \cref{eq:pdsg_calibrated}. As shown in \cref{fig:pdsg}(c), the distribution of gradients is balanced, ensuring that the gradients on both sides of the threshold receive equal attention. In this paper, we set $b = 0.5\boldsymbol{\sigma}$ in all experiments, which will be discussed in \cref{sec:ablation}.
\begin{equation}
    \frac{\partial \boldsymbol{s}^{l}[t]}{\partial \boldsymbol{u}^{l}[t]} = \frac{1}{\sqrt{2\pi}\boldsymbol{\sigma}}\text{exp}(-\frac{(\boldsymbol{u}^{l}[t]-V_{th}-b)^2}{2\boldsymbol{\sigma}^2})
    \label{eq:pdsg_calibrated}
\end{equation}

\subsection{Sparse Dynamic Attack}
\begin{figure*}[tb]
  \centering
  \includegraphics[height=8cm]{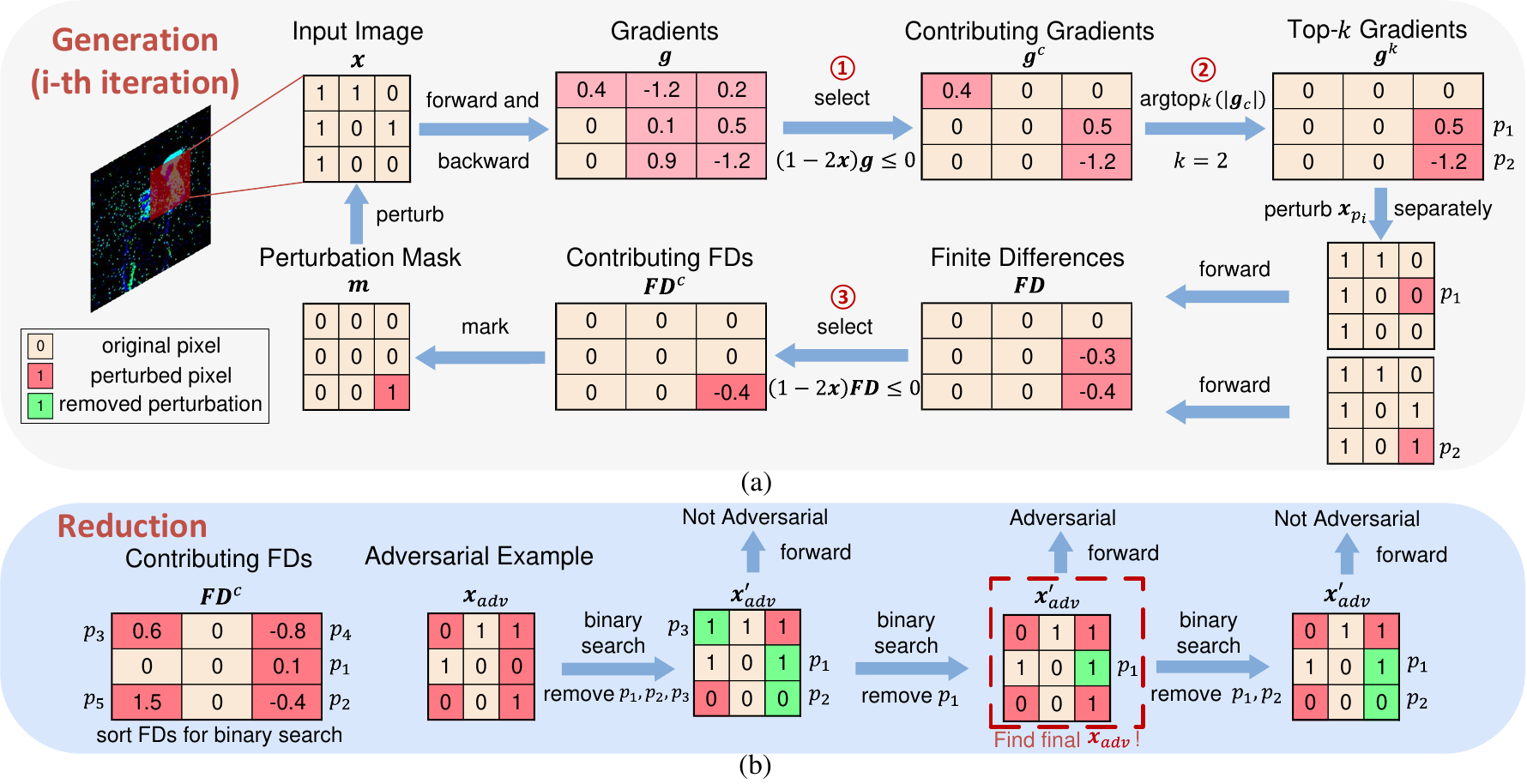}
  \caption{Illustration of our sparse dynamic attack. (a) In the generation process, we select contributing gradients through their signs, achieve top-$k$ significant gradients and calculate their FDs to add perturbations. (b) In the reduction process, we sort the contributing FDs calculated by the generation process, then adopt binary search to find $p_2$ which makes the example cease to be adversarial after removed.}
  \label{fig:sda}
\end{figure*}

Gradient-based adversarial attacks are not applicable to binary dynamic images since these attacks produce floating-point gradients. To effectively generate sparse perturbations, we propose a generation-reduction paradigm for our SDA. The generation process combines the gradients and finite differences (FDs) to select the most desirable pixels to attack. The reduction process leverages recorded FDs to remove redundant perturbations.

\subsubsection{Generation}
{\bf{\circled{1}Contributing Gradients.}} Since the gradients are useful for rapid estimation, we first calculate the gradients of the input through STBP and our aforementioned PDSG as below:
\begin{equation}
    \boldsymbol{g} = \frac{\partial \mathcal{L}(f(\boldsymbol{x}),y)}{\partial \boldsymbol{x}}.
    \label{eq:gradient}
\end{equation}

To prevent gradient vanishing, we adopt the loss function from C\&W \cite{carlini2017towards}:
\begin{equation}
    \mathcal{L}(f(\boldsymbol{x}),y)=\text{max}\{f(\boldsymbol{x})_y-\text{max}_{i\neq y}{f(\boldsymbol{x})_i},0\}.
    \label{eq:cwloss}
\end{equation}

When the loss decreases to zero, the attack will be successful. According to \cref{eq:gradient}, the change of the input $\Delta \boldsymbol{x}$ contributes to the decline of the loss only when $\Delta \boldsymbol{x}$ and $\boldsymbol{g}$ have opposite signs. As the change of the input is unique for binary inputs ($\Delta x_i = 1$ for $x_i=0$ and $\Delta x_i = -1$ for $x_i=1$), we select the pixels which have contributing gradients and are not in the perturbation mask $m$:
\begin{equation}
    \boldsymbol{g}^c = \boldsymbol{g} \cdot ((1-2\boldsymbol{x})\cdot \boldsymbol{g} <= 0) \cdot (1 - \boldsymbol{m}).
    \label{eq:contributing_g}
\end{equation}

{\bf{\circled{2}Top-$k$ gradients.}} To optimize the sparsity of perturbations, we select only a portion of pixels for further calculations. As the pixels with larger gradient values are considered to contribute more adversary \cite{dong2020greedyfool, croce2022sparse}, we select $k$ pixels with the largest absolute gradient values, where $k$ is incremented by $k_{init}$ for each iteration. For all experiments, we set $k_{init}=10$. In the $n$-th iteration, the top-$k$ gradients are obtained as:
\begin{equation}
    \boldsymbol{g}^k = \{ g^c_i | i \in \text{argtop}k (|\boldsymbol{g}_c|), k=(n+1)k_{init}\}.
    \label{eq:topk_g}
\end{equation}

{\bf{\circled{3}Contributing FDs.}} The process of selecting top-$k$ gradients is coarse. The gradients are inaccurate because the change of the input ($\Delta \boldsymbol{x} = \pm 1$) is large compared to the infinitesimal in \cref{eq:gradient}. To perform a fine-grained selection, we leverage the finite difference, which reflects the actual trend of the loss when a binarized pixel is perturbed:
\begin{equation}
    FD_i(\boldsymbol{x}) = \frac{\mathcal{L}(f(\boldsymbol{x} + \Delta x_i\boldsymbol{e}_i),y) - \mathcal{L}(f(\boldsymbol{x}),y)}{\Delta x_i}\boldsymbol{e}_i.
    \label{eq:fd}
\end{equation}

In \cref{eq:fd}, $\Delta x_i$ denotes the change in the $i$-th index of the input, and $\boldsymbol{e}_i$ is the standard basis vector with 1 at $i$-th index. $FD_i(\boldsymbol{x})$ represents the change of the loss when the $i$-th index of the input is perturbed, which approximates to the gradient when $\Delta x_i \rightarrow 0$. We only calculate the FDs for the pixels selected by top-$k$ gradients, and the forward process can be parallelized for acceleration. Similar to contributing gradients, we select contributing FDs for pixels whose change and FD have opposite signs:
\begin{equation}
    \boldsymbol{FD}^c = \boldsymbol{FD} \cdot ((1-2\boldsymbol{x})\cdot \boldsymbol{FD} <= 0).
    \label{eq:contributing_fd}
\end{equation}

Finally, we add the pixels with valid $FD^c$ to the perturbation mask $\boldsymbol{m}$, perturb the input through the mask in the current iteration, and iteratively perform the generation process until the input becomes adversarial. The entire generation process is illustrated in \cref{fig:sda}(a).

\subsubsection{Reduction}
In the generation process, we have already obtained an adversarial example $\boldsymbol{x}_{adv}$ with sparse perturbations. Nonetheless, the selection of the top-$k$ gradients leads to a local optimum. Therefore, we devise a perturbation reduction method to eliminate perturbations with minor impact. 

Intuitively, we believe the pixel with the smallest absolute value of the FD has the least influence on the loss, and removing the pixel with the smallest FD is unlikely to affect the classification result. Consequently, we attempt to eliminate perturbations sequentially in ascending order of their absolute FDs. As shown in \cref{fig:sda}(b), we construct a sorted set of perturbed pixels, where $n$ is the number of perturbed pixels:
\begin{equation}
    \mathcal{S} = \{p_i | \  |FD^c_{p_1}|<\cdots<|FD^c_{p_n}|, \ i=1,\cdots,n\}.
    \label{eq:sort}
\end{equation}

Then we sequentially remove the perturbations $p_i$ from $p_1$ to $p_n$. If $\boldsymbol{x}_{adv}$ ceases to be adversarial upon removing $p_j$, we consider $\mathcal{S}_1=\{p_1,\cdots,p_{j-1}\}$ as redundant perturbations which have no affect on the classification result, and $\mathcal{S}_2=\{p_j,\cdots,p_n\}$ as necessary perturbations. Therefore, the objective of the reduction process is to find $j$ in a sorted set. To improve the efficiency, we adopt the binary search to reduce the complexity to $O(\text{log}\ n)$. An example with $j=2$ and $n=5$ is illustrated in \cref{fig:sda}(b). In general, this reduction process removes dispensable perturbations and fully optimizes the sparsity of perturbations.
\section{Experiments}

\subsection{Experimental Setup}
\begin{table*}[tb]
  \centering
  \setlength\tabcolsep{6.5pt}
  \footnotesize
  \begin{tabular}{ccccccccccccc}
    \hline
    \multirow{3}{*}{Dataset} & \multirow{3}{*}{Architecture} & \multirow{3}{*}{Input} & \multirow{3}{*}{\makecell{Acc.\\(\%)}}& \multirow{3}{*}{Attack} & \multicolumn{4}{c}{ASR. (\%) ($\epsilon=2 / 255$)} & \multicolumn{4}{c}{ASR. (\%) ($\epsilon=8 / 255$)}\\
    \cline{6-13}
    & & & & & \multirow{2}{*}{STBP} & \multirow{2}{*}{RGA} & \multirow{2}{*}{HART} & \multirow{2}{*}{\makecell{{\bf{PDSG}}\\{\bf{(Ours)}}}} & \multirow{2}{*}{STBP} & \multirow{2}{*}{RGA} & \multirow{2}{*}{HART} & \multirow{2}{*}{\makecell{{\bf{PDSG}}\\{\bf{(Ours)}}}} \\
    & & & & & & & & & & & & \\
    \hline
    \multirow{8}{*}{CIFAR10} & \multirow{2}{*}{ResNet18} & \multirow{2}{*}{Direct} & \multirow{2}{*}{94.72} & FGSM & 38.21 & 31.14 & 37.30 & \textbf{43.98} & 52.36 & 45.80 & 46.72 & \textbf{79.56}\\
    \cline{5-13}
    & & & & PGD & 66.74 & 61.97 & 66.77 & \textbf{69.62} & 99.81 & 92.47 & 98.64 & \textbf{100.00}\\
    \cline{2-13}
    & \multirow{2}{*}{\makecell{ResNet18 \\ (Adv. trained)}} & \multirow{2}{*}{Direct} & \multirow{2}{*}{90.65} & FGSM & 4.10 & 6.66 & 8.66 & \textbf{9.29} & 22.31 & 34.64 & 47.18 & \textbf{47.42}\\
    \cline{5-13}
    & & & & PGD & 4.21 & 7.56 & 9.95 & \textbf{10.68} & 28.23 & 46.50 & 59.56 & \textbf{62.16}\\
    \cline{2-13}
    & \multirow{2}{*}{ResNet18} & \multirow{2}{*}{Poisson} & \multirow{2}{*}{76.98} & FGSM & 5.40 & 5.85 & 6.64 & \textbf{7.05} & 28.92 & 25.58 & 31.57 & \textbf{32.59}\\
    \cline{5-13}
    & & & & PGD & 8.39 & 7.63 & \textbf{8.50} & 8.22 & 39.66 & 35.78 & \textbf{42.30} & 39.98\\
    \cline{2-13}
    & \multirow{2}{*}{VGG11} & \multirow{2}{*}{Direct} & \multirow{2}{*}{94.08} & FGSM & 26.93 & 21.39 & 27.29 & \textbf{30.45} & 45.37 & 37.02 & 38.89 & \textbf{82.71}\\
    \cline{5-13}
    & & & & PGD & 41.07 & 39.07 & \textbf{48.12} & 39.20 & 98.35 & 85.67 & 97.35 & \textbf{99.71}\\
    \hline
    \multirow{2}{*}{CIFAR100} & \multirow{2}{*}{ResNet18} & \multirow{2}{*}{Direct} & \multirow{2}{*}{75.94} & FGSM & 56.16 & 52.45 & 59.01 & \textbf{59.05} & 71.50 & 64.43 & 70.32 & \textbf{83.29}\\
    \cline{5-13}
    & & & & PGD & 81.92 & 78.36 & \textbf{86.95} & 78.50 & 99.60 & 98.00 & 99.62 & \textbf{99.83}\\
    \hline
    \multirow{2}{*}{ImageNet} & \multirow{2}{*}{HST-10-768} & \multirow{2}{*}{Direct} & \multirow{2}{*}{84.28} & FGSM & 47.52 & 0.47 & \textbf{57.12} & 54.89 & 56.29 & 4.97 & 76.61 & \textbf{79.71}\\
    \cline{5-13}
    & & & & PGD & 90.01 & 0.03 & \textbf{91.41} & 87.43 & 99.89 & 2.84 & 99.98 & \textbf{100.00}\\
    \hline
    \multirow{2}{*}{CIFAR10DVS} & \multirow{2}{*}{VGGSNN} & \multirow{2}{*}{Integer} & \multirow{2}{*}{78.80} & FGSM & \textbf{9.90} & 8.25 & 9.52 & 9.64 & 41.37 & 35.28 & 42.26 & \textbf{42.39}\\
    \cline{5-13}
    & & & & PGD & 9.39 & 7.74 & 10.15 & \textbf{10.15} & \textbf{45.94} & 37.31 & 45.81 & 44.54\\
    \hline
  \end{tabular}
  \caption{Comparison with state-of-the-art approaches on attacking static images and integer dynamic frames. ASR. denotes the attack success rate. $\epsilon$ is the attack intensity. STBP denotes attacking using training-phase SG. The best results are in bold.}
  \label{tab:static}
\end{table*}
\begin{table*}[tb]
  \centering
  \setlength\tabcolsep{3.6pt}
  \footnotesize
  \begin{tabular}{cccccccccccccccc}
    \hline
    \multirow{3}{*}{Dataset} & \multirow{3}{*}{Architecture} & \multirow{3}{*}{Input} & \multirow{3}{*}{\makecell{Acc.\\(\%)}}& \multirow{3}{*}{Attack} & \multicolumn{4}{c}{ASR. (\%)  ($\ell_0 < 200$)} & \multicolumn{4}{c}{ASR. (\%)  ($\ell_0 < 800$)}& \multicolumn{3}{c}{Dynamic Evaluation}\\
    \cline{6-16}
    & & & & & \multirow{2}{*}{STBP} & \multirow{2}{*}{RGA} & \multirow{2}{*}{HART} & \multirow{2}{*}{\makecell{{\bf{PDSG}}\\{\bf{(Ours)}}}} & \multirow{2}{*}{STBP} & \multirow{2}{*}{RGA} & \multirow{2}{*}{HART} & \multirow{2}{*}{\makecell{{\bf{PDSG}}\\{\bf{(Ours)}}}} & \multirow{2}{*}{ASR.} & \multirow{2}{*}{\makecell{Mean\\$\ell_0$}} & \multirow{2}{*}{\makecell{Median\\$\ell_0$}}\\
    & & & & & & & & & & & & & & \\
    \hline
    \multirow{4}{*}{\makecell{N-MNIST}} & \multirow{4}{*}{PLIFNet} & \multirow{4}{*}{Binary} & \multirow{4}{*}{99.57} & SCG & 0.0 & 0.0 & 0.0 & 0.0 & 18.0 & 15.0 & 0.0 & 34.0 & 91.0 & 1144.90 & 1162.00 \\
    \cline{5-16}
    & & & & SpikeFool & 15.0 & 0.0 & 2.0 & 1.0 & 93.0 & 20.0 & 46.0 & 27.0 & 97.0 & 444.44 & 356.00\\
    \cline{5-16}
    & & & & GSAttack & 0.0 & 0.0 & 0.0 & 0.0 & 0.0 & 0.0 & 0.0 & 0.0 & 91.0 & 2828.25 & 2905.00 \\
    \cline{5-16}
    & & & & \textbf{SDA(Ours)} & 23.0 & 9.0 & 22.0 & \textbf{63.0} & 93.0 & 86.0 & 92.0 & \textbf{99.0} & \textbf{100.0} & \textbf{207.34} & \textbf{171.00} \\
    \hline
    \multirow{4}{*}{\makecell{DVS-\\Gesture}} & \multirow{4}{*}{VGGSNN} & \multirow{4}{*}{Binary} & \multirow{4}{*}{95.14} & SCG & 0.0 & 0.0 & 0.0 & 0.0 & 2.0 & 0.0 & 0.0 & 0.0 & 100.0 & 8377.84 & 7586.50\\
    \cline{5-16}
    & & & & SpikeFool & 3.0 & 2.0 & 1.0 & 2.0 & 16.0 & 12.0 & 6.0 & 14.0 & 69.0 & 2762.41 & 1908.00\\
    \cline{5-16}
    & & & & GSAttack & 0.0 & 0.0 & 0.0 & 0.0 & 0.0 & 0.0 & 0.0 & 0.0 & 71.0 & 9820.14 & 8521.50\\
    \cline{5-16}
    & & & & \textbf{SDA(Ours)} & 19.0 & 10.0 & 16.0 & \textbf{21.0} & 38.0 & 39.0 & 40.0 & \textbf{52.0} & \textbf{100.0} & \textbf{1731.63} & \textbf{769.50}\\
    \hline
    \multirow{4}{*}{\makecell{CIFAR10\\-DVS}} & \multirow{4}{*}{ResNet18} & \multirow{4}{*}{Binary} & \multirow{4}{*}{78.20} & SCG & 0.0 & 0.0 & 0.0 & 0.0 & 4.0 & 0.0 & 0.0 & 0.0 & 100.0 & 2346.10 & 2191.00\\
    \cline{5-16}
    & & & & SpikeFool & 19.0 & 2.0 & 13.0 & 0.0 & 70.0 & 13.0 & 33.0 & 8.0 & 100.0 & 674.89 & 491.00\\
    \cline{5-16}
    & & & & GSAttack & 0.0 & 0.0 & 0.0 & 0.0 & 0.0 & 0.0 & 0.0 & 0.0 & 65.0 & 8511.60 & 8741.00\\
    \cline{5-16}
    & & & & \textbf{SDA(Ours)} & 34.0 & 12.0 & 21.0 & \textbf{38.0} & 78.0 & 34.0 & 57.0 & \textbf{82.0} & \textbf{100.0} & \textbf{458.02} & \textbf{303.00}\\
    \hline
  \end{tabular}
  \caption{Comparison with state-of-the-art approaches on SNN-based binary attack. $\ell_0 < 200$ means the number of modified pixels is less than $200$. We incorporate the PDSG into our SDA in the dynamic evaluation. The best results are in bold.}
  \label{tab:dynamic}
\end{table*}
We validate the effectiveness of our PDSG on both static and dynamic datasets, and our SDA on dynamic datasets. CIFAR10/100 \cite{krizhevsky2009learning} and ImageNet \cite{deng2009imagenet} are adopted as static datasets, while N-MNIST \cite{orchard2015converting}, DVS-Gesture \cite{amir2017low} and CIFAR10-DVS \cite{li2017cifar10} are utilized as dynamic datasets. As the dynamic datasets are all in event-stream forms, we utilize SpikingJelly \cite{fang2023spikingjelly} framework to aggregate the events into 10 frames, and binarize the frames by capping the maximum value of each pixel to $1$ \cite{he2020comparing, buchel2022adversarial}. The input size for N-MNIST is $34\times34$, and for DVS-Gesture and CIFAR10-DVS it is $128\times128$, aligning with their original input sizes. The models contain spiking ResNet-18 \cite{huang2024clif}, spiking VGG-11 \cite{bu2023rate}, VGGSNN \cite{deng2022temporal}, and hierarchical spiking transformer (HST) \cite{zhou2024qkformer}. The timestep for static datasets is 4 and for dynamic datasets is 10. Details of the datasets and models are provided in the Appendix.

The evaluation metrics for the experiments include the attack success rate (ASR) and the $\ell_0$-norm of perturbations. For attacking binary inputs, we randomly select 100 correctly classified inputs in the test set to perform attacks, and the attack fails when iterations exceed 500.

\subsection{Comparison with State-of-the-art Works}

In this section, we demonstrate the effectiveness of our PDSG and SDA by comparing them with state-of-the-art (SOTA) adversarial attacks on SNNs. We compare our PDSG with RGA \cite{bu2023rate} and HART \cite{hao2024threaten}, which focus on optimizing the gradient flow and adopt universal SGs, and compare our SDA with SCG \cite{liang2021exploring}, SpikeFool \cite{buchel2022adversarial}, and GSAttack \cite{yao2024exploring}, which perform attacks on binary dynamic images.

The results for attacking static images and integer dynamic frames in various attack intensities are shown in \cref{tab:static}. On CIFAR10, we use ResNet18 with standard training and adversarial training. The adversarial training is conducted by PGD attack with $\epsilon=2/255$ \cite{rice2020overfitting}. For ResNet18 in standard training, our PDSG significantly surpasses SOTA methods and STBP (attacking using the training-phase SG). The results also indicate that the training-phase SG is not always the most effective during attack. Although adversarial training effectively reduces the ASR, our PDSG is the least affected by the defense. For other models and datasets, our PDSG performs the most stably and has the ability to maintain high ASR in various scenarios. Notably on ImageNet, the PDSG achieves 100\% ASR.

For evaluating our SDA on attacking binary dynamic images, in \cref{tab:dynamic}, we measure the ASR under $\ell_0$-norm bounded attack and unbounded attack. As SCG and GSAttack do not specifically optimize the sparsity of perturbations, their $\ell_0$ are insufficient. Our SDA completely outperforms SpikeFool in terms of attack success rate and the sparsity. When combined with the PDSG, our SDA achieves 82\% ASR under a constraint of $\ell_0 < 800$ on CIFAR10DVS, which is only $0.24\%$ of the input pixels. In dynamic evaluation, the SDA perturbs a median of 303 pixels on CIFAR10DVS, which is only 62\% of the SpikeFool.

\subsection{Effects of the PDSG}
Our PDSG focuses on addressing the issue of invisible SGs during attacks on SNNs. To demonstrate the adaptability of our PDSG, we first train ResNet18 on CIFAR10 using various shapes and parameters of SGs, with further details shown in the Appendix. Then we adopt FGSM to attack these models with various attack-phase SGs. The results are presented in \cref{tab:pdsg}. We observe that the ASR is highly dependent on the attack-phase SG. For fixed SGs, although the ATan SG performs well in most experiments, it is defeated by the Rect.(2) SG when the model is also trained using the Rect.(2) SG. Therefore, it is challenging to identify a fixed SG that consistently exhibits stable attack performance. Our PDSG demonstrates the best performance across all experiments, achieving approximate 80\% ASR. 

\begin{table}[tb]
  \centering
  \setlength\tabcolsep{2.6pt}
  \footnotesize
  \begin{tabular}{@{}cccccccc@{}}
    \hline
    \multirow{2}{*}{\makecell{Training\\SG}} & \multirow{2}{*}{\makecell{Acc.\\(\%)}} & \multicolumn{6}{c}{Attack SG} \\
    \cline{3-8}
    & & Rect.(1) & Rect.(2) & Rect.(0.5) & ATan & Triangle & PDSG \\
    \hline
    Rect.(1) & 94.72 & 52.36 & 64.13 & 27.20 & 68.56 & 41.29 & \textbf{79.56} \\
    Rect.(2) & 93.25 & 48.33 & 73.03 & 20.99 & 59.41 & 37.06 & \textbf{84.68}\\
    Rect.(0.5) & 94.58 & 44.50 & 50.73 & 29.25 & 65.35 & 38.41 & \textbf{79.85} \\
    ATan & 94.67 & 47.10 & 60.98 & 25.72 & 66.91 & 37.79 & \textbf{82.07} \\
    Triangle & 95.01 & 44.90 & 55.69 & 26.34 & 68.09 & 36.31 & \textbf{79.77} \\
    PDSG & 90.69 & 60.57 & 65.99 & 26.76 & 64.73 & 51.44 & \textbf{82.02}\\
  \hline
  \end{tabular}
  \caption{Attack success rate of FGSM ($\epsilon=8/255$) on ResNet18 in cases of various training SGs and attack SGs. {\it Rect.}(1) means using the Rectangular SG with $w=1$. The best results are in bold.}
  \label{tab:pdsg}
\end{table}

\subsection{Ablation study}
\label{sec:ablation}
\begin{figure}[tb]
  \centering
  \includegraphics[height=3.1cm]{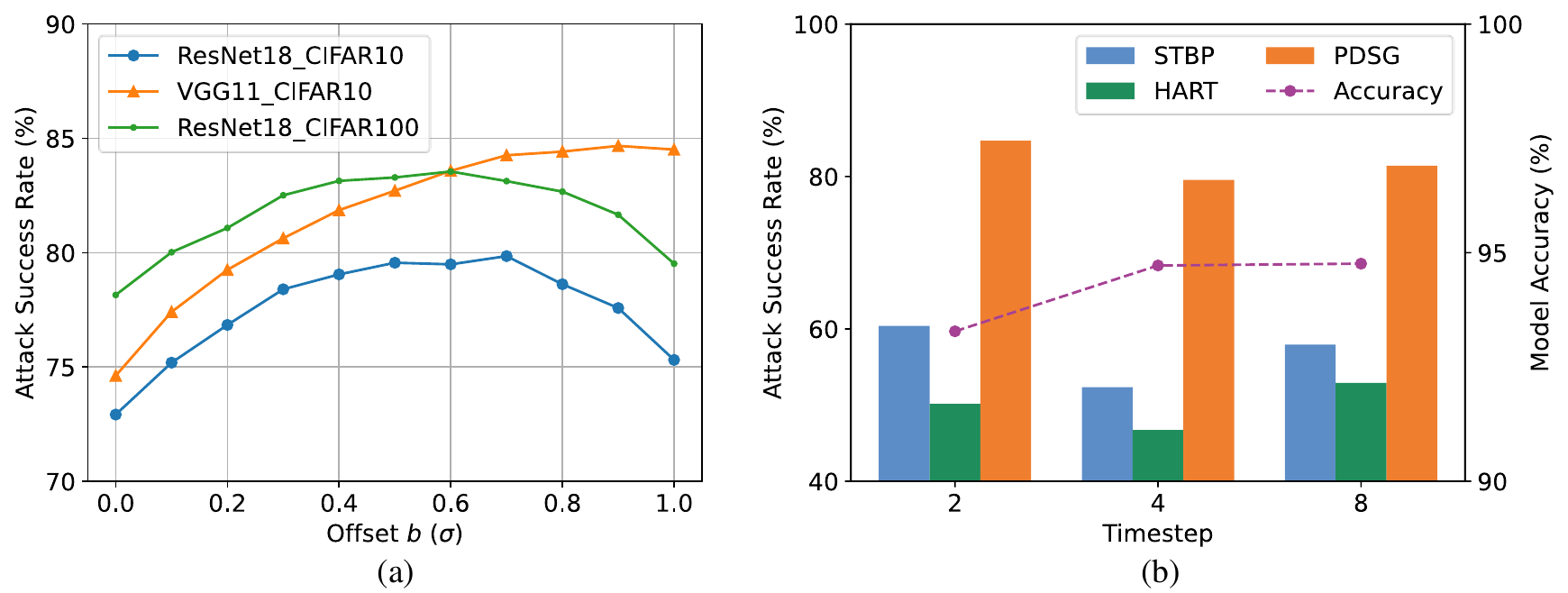}
  \caption{(a) Effectiveness of various offsets in the calibration of our PDSG. We adopt FGSM ($\epsilon=8/255$) to perform attacks. (b) Impact of diverse timesteps in attacking ResNet18 on CIFAR10.}
  \label{fig:ablation}
\end{figure}

We first evaluate the effectiveness of the offset $b$ in our PDSG. As illustrated in \cref{fig:ablation}(a), the performance of the attack is sub-optimal before calibration ($b=0$) due to the imbalanced distribution of membrane potential. After calibration, the ASR significantly improves as $b$ increases within an appropriate range. However, the ASR will also decrease when $b$ is excessively large. Therefore, to achieve stable performance, we adopt $b = 0.5\sigma$ in all experiments. Moreover, we discuss the impact of diverse timesteps of the model. As shown in \cref{fig:ablation}(b), the accuracy increases with the timestep; however, the ASR is independent from the timestep, indicating that the performance of the attack is related to the distribution of the actual model. Our PDSG performs the best at all timesteps, demonstrating that the PDSG can adapt to various timesteps of the model.

\cref{tab:ablation_component} shows that every optimization component plays an important role in our SDA. We set the top-$k$ gradients in \cref{eq:topk_g} as the base component because it provides a basic sparse selection of perturbations. When the C\&W loss in \cref{eq:cwloss} is adopted, although the mean $\ell_0$ slightly increased, the ASR is improved due to resolved gradient vanishing. After introducing the FDs in \cref{eq:fd} and the reduction process in \cref{eq:sort}, redundant perturbations are effectively removed, jointly improving the sparsity and the ASR. Cooperating with the PDSG, the representation of the gradients is fully optimized, thereby demonstrating superior attack performance of our SDA.
\begin{table}[tb]
  \centering
    \setlength\tabcolsep{4pt}
    \footnotesize
    \begin{tabular}{@{}cccccccc@{}}
    \hline
    \makecell{Top$k$ \\ gradients} & \makecell{C\&W\\Loss} & FDs & Reduction & PDSG & \makecell{ASR. \\ (\%)} & \makecell{Mean \\ $\ell_0$}  & \makecell{Median \\ $\ell_0$}\\
    \hline
    \checkmark & & & & & 69.0 & 655.70 & 450.0\\
    \checkmark & \checkmark & & & & 72.0 & 678.90 & 450.0\\
    \checkmark & \checkmark & \checkmark & & & 75.0 & 567.82 & 389.5\\
    \checkmark & \checkmark & \checkmark & \checkmark & & 78.0 & 521.26 & 351.5\\
    \checkmark & \checkmark & \checkmark & \checkmark & \checkmark & {\bf 82.0} & {\bf 458.02} & {\bf 303.0}\\
    \hline
    \end{tabular}
    \caption{The effect of each component in our SDA method on CIFAR10DVS dataset. \checkmark denotes utilizing the component. The best results are in bold.}
    \label{tab:ablation_component}
\end{table}

\section{Conclusion}
In this paper, we introduce the potential-dependent surrogate gradient to adaptively address invisible SGs in attacking SNNs. Moreover, a novel sparse dynamic attack method is proposed to effectively attack binary dynamic images on SNNs with sparse perturbations. The experiments demonstrate that our PDSG and SDA achieve superior performance with 100\% attack success rate on ImageNet and 82\% attack success rate by modifying only 0.24\% of the pixels on CIFAR10DVS. Looking ahead, we hope that our proposed methods will inspire further research to comprehensively rethink the robustness of SNNs in various scenarios and to redesign corresponding defense mechanisms.

\section*{Acknowledgment}
This paper is supported by the National Natural Science Foundation of China (NSF) under Grant No.\@ 62406008.
{
    \small
    \bibliographystyle{ieeenat_fullname}
    \bibliography{main}
}
\clearpage
\setcounter{page}{1}
\setcounter{section}{0}
\setcounter{figure}{0}
\setcounter{equation}{0}
\setcounter{table}{0}
\renewcommand{\thesection}{S\arabic{section}}
\renewcommand{\thefigure}{S\arabic{figure}}
\renewcommand{\thetable}{S\arabic{table}}
\renewcommand{\theequation}{S\arabic{equation}}
\maketitlesupplementary

\section{Derivation of Potential-Dependent Surrogate Gradient}

We adopt the two-point zeroth-order method to calculate the gradient of the firing function approximately \cite{mukhoty2023direct}:
\begin{equation}
    G^2(u;z,\delta)=\frac{h(u+z\delta-V_{th}) - h(u-z\delta-V_{th})}{2\delta}z.
    \label{eq:zeroth_order_sup}
\end{equation}

Due to the firing function defined in Eq. (2) of the main text, the two-point zeroth-order can be simplified as:
\begin{equation}
    G^2(u;z,\delta)=
    \begin{cases}
        \frac{|z|}{2\delta}, &|u-V_{th}|<|z\delta |\\
        0, &\text{otherwise}\\
    \end{cases}.
    \label{eq:zeroth_order_cases_sup}
\end{equation}

Since $z$ is sampled from the distribution $\lambda$, the surrogate gradient equals to the expectation of the two-point zeroth-order \cite{liu2020primer}:
\begin{align}
    \frac{\partial s}{\partial u}
    &=\mathbb{E}_{z \sim \lambda}[G^2(u;z,\delta)] \notag \\
    &=\int_{-\infty}^{+\infty} G^2(u;z,\delta)\lambda(z)dz \notag \\
    &=2\int_{\frac{|u-V_{th}|}{\delta}}^{\infty} \frac{|z|}{2\delta}\lambda(z)dz \notag \\
    &=\int_{\frac{|u-V_{th}|}{\delta}}^{\infty} \frac{|z|}{\delta}\lambda(z)dz.
    \label{eq:surrogate_sup}
\end{align}

As demonstrated in \cref{sec:pdsg}, $u + z\delta$ follows a normal distribution $\mathcal{N}(\mu, \sigma^2)$, where $\mu$ denotes the mean of membrane potential, and $\sigma$ is the standard deviation of the membrane potential. Therefore, $z\sim \mathcal{N}(\frac{\mu-u}{\delta}, \frac{\sigma^2}{\delta^2})$. Following the requirement $z\sim\mathcal{N}(0,1)$ in the two-point zeroth-order method, we set $\delta=\sigma$, and when $u \approx \mu$, we get:
\begin{align}
    \frac{\partial s}{\partial u}
    &=\int_{\frac{|u-V_{th}|}{\boldsymbol{\sigma}}}^{\infty} \frac{|z|}{\boldsymbol{\sigma}}\cdot \frac{1}{\sqrt{2\pi}} \text{exp}(-\frac{z^2}{2})dz \notag \\
    &=\frac{1}{\sqrt{2\pi}\boldsymbol{\sigma}}\int_{\frac{|u-V_{th}|}{\boldsymbol{\sigma}}}^{\infty} \text{exp}(-\frac{z^2}{2})d(\frac{z^2}{2}) \notag \\
    &=\frac{1}{\sqrt{2\pi}\boldsymbol{\sigma}}\text{exp}(-\frac{(\boldsymbol{u}^{l}[t]-V_{th})^2}{2\boldsymbol{\sigma}^2}).
    \label{eq:pdsg_sup}
\end{align}

Here, we follow the TAB \cite{jiang2024tab} to adopt the temporal accumulated channel-wise standard deviation $\boldsymbol{\sigma}$ of membrane potential.

\section{Algorithm of Sparse Dynamic Attack}

\begin{algorithm}  
\caption{Sparse Dynamic Attack (SDA)}  
\label{alg:sda}
{\bf Input:} Classifier $f$, benign image $\boldsymbol{x}$, label $y$.\\
{\bf Parameters:} Initial gradient selection count $k_{init}$, \\maximum number of iterations $N$. \\
{\bf Output:} Adversarial example $\boldsymbol{x}_{adv}$.
\begin{algorithmic}[1]

\State \textbf{\#Generation Process:}
\State Initialize perturbation mask $\boldsymbol{m} \leftarrow 0$ 
\State Initialize contributing FDs $\boldsymbol{FD}^c \leftarrow \infty$
\State Initialize $\boldsymbol{x}^0 \leftarrow \boldsymbol{x}$
\For{$n = 0$ to $N-1$}
    \State \text{Calculate the gradient }$\boldsymbol{g}(\boldsymbol{x}^n)$ \Comment{\cref{eq:gradient}}
    \State $\boldsymbol{g}^c \leftarrow \boldsymbol{g} \cdot ((1-2\boldsymbol{x}^n)\cdot \boldsymbol{g} <= 0) \cdot (1 - \boldsymbol{m})$ \Comment{\cref{eq:contributing_g}}
    \State $k \leftarrow (n+1) k_{init}$
    \State $p_1, p_2, \dots , p_k \leftarrow \text{argtop}k(|\boldsymbol{g}^c|)$ \Comment{\cref{eq:topk_g}}
    \For{$i=1$ to $k$} \Comment{Parallelized}
        \State \text{Calculate } $FD_{p_i}(\boldsymbol{x}^n)$ \Comment{\cref{eq:fd}}
        \If{$(1-2x^n_{p_i}) \cdot FD_{p_i} <= 0$} \Comment{\cref{eq:contributing_fd}}
            \State $FD^c_{p_i} \leftarrow FD_{p_i}$
            \State $m_{p_i} \leftarrow 1$
        \EndIf
    \EndFor
    \State $\boldsymbol{x}^{n+1} \leftarrow \boldsymbol{x} \cdot (1-\boldsymbol{m}) + (1-\boldsymbol{x}) \cdot \boldsymbol{m}$ \Comment{Perturb} 
    \If{$\boldsymbol{x}^{n+1}$ is adversarial}
        \State $\boldsymbol{x}_{adv} \leftarrow \boldsymbol{x}^{n+1}$
        \State \textbf{break}  
    \EndIf
    \If{$n == N-1$}
        \State \textbf{Attack failed}
    \EndIf
\EndFor
\State \textbf{\#Reduction Process:}
\State Construct sorted perturbed indices $\mathcal{S}$ \Comment{\cref{eq:sort}}
\State Initialize $L\leftarrow0, R\leftarrow \text{len}(\mathcal{S)}-1$
\While {$L<=R$} 
    \State $j \leftarrow \lfloor \frac{L+R}{2} \rfloor$
    \State $\boldsymbol{x}_{adv}'\leftarrow \boldsymbol{x}_{adv}$
    \State $\boldsymbol{x}_{adv}'[\mathcal{S}[0:j+1]]\leftarrow 1-\boldsymbol{x}_{adv}'[\mathcal{S}[0:j+1]]$
    \If{$\boldsymbol{x}'_{adv}$ is adversarial}
        \State $L \leftarrow j+1$
        \State $\boldsymbol{x}_{final} \leftarrow \boldsymbol{x}'_{adv}$
    \Else
        \State $R \leftarrow j-1$
    \EndIf
\EndWhile
\State \textbf{return } final adversarial example $\boldsymbol{x}_{final}$
\end{algorithmic}  
\end{algorithm}
\clearpage

\section{Adversarial Threat Model}

\begin{figure}[tb]
  \centering
  \includegraphics[width=\linewidth]{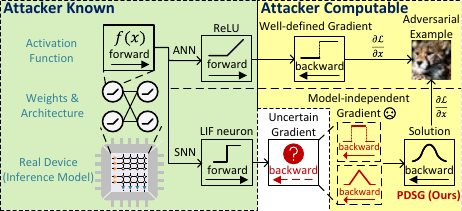}
   \caption{White-box attacker knows weights and architecture of the model, which is enough for attacking ANNs. For SNNs, the gradient of activation is not exposed during the inference. Our PDSG provides a solution for the problem of uncertain gradient.}
   \label{fig:rebut_illustration}
\end{figure}

As shown in \cref{fig:rebut_illustration}, in white-box attacks, the attacker leverages gradients to perform attacks. As the activation in ANNs has a well-defined gradient, the gradients can be directly calculated through the model weights and architecture, and the attacker does not require training details, which are useless for attack.

In contrast, the activation in SNNs does not have exact backward function. During the training stage, the surrogate gradient is adopted as the backward function. However, the inference model does not store the backward function used during training; further, as shown in \cref{tab:pdsg}, adopting it for attack does not guarantee the performance. Therefore, the {\bf invisible surrogate gradients} means: the backward function in training stage is invisible during inference and attack, and the optimal backward function is uncertain. 

In summary, the adversarial threat model in our paper is identical to white-box ANN attacks, which is: {\bf the attacker knows the weights, architecture, and the activation's forward function of the victim inference model.} The backward function is inaccessible. This adversarial threat model is suitable for real-world situations: {\bf the attacker obtains a neuromorphic device, where the backward function is served as a training skill and not stored in the device.} Instead of adopting model-independent backward functions \cite{bu2023rate, hao2024threaten}, our adaptive PDSG effectively increases the attack success rate.

\section{Details of Experiments}

{\bf{Details of datasets. }}CIFAR10/100 \cite{krizhevsky2009learning} dataset contains 60,000 images with 10/100 classes, which are split into the training set with 50,000 images and test set with 10,000 images. The input size is 32$\times$32.

ImageNet \cite{deng2009imagenet} dataset contains 1,281,167 images as training set and 50,000 images as validation set. The number of classes is 1000, and the input size is 224$\times$224.

NMNIST \cite{orchard2015converting} dataset is constructed by saccading the MNIST dataset \cite{lecun1998gradient} using DVS. The training set contains 60000 samples, and the test set contains 10000 samples. The size of frames is 34$\times$34.

DVS-Gesture \cite{amir2017low} dataset includes samples of hand gesture recorded by DVS128 camera. The training set contains 1176 samples, and the test set contains 288 samples. The size of frames is 128$\times$128.

CIFAR10-DVS \cite{li2017cifar10} dataset is converted from CIFAR10 \cite{krizhevsky2009learning} dataset, including 10000 samples with 10 classes. We split these 10000 samples into 9000 training samples and 1000 test samples. The size of frames is 128$\times$128.

{\bf{Details of models. }} We adopt spiking ResNet-18 \cite{huang2024clif}, spiking VGG-11 \cite{bu2023rate}, VGGSNN \cite{deng2022temporal}, PLIFNet \cite{fang2021incorporating}, and hierarchical spiking transformer (HST) \cite{zhou2024qkformer} in our experiments. The spiking ResNet-18 and spiking VGG-11 maintain the same architecture as the original ResNet-18 \cite{he2016deep} and VGG-11 \cite{simonyan2014very}, respectively, with the activation function replaced by LIF neurons. The VGGSNN removes the last two linear layers of the spiking VGG-11. The PLIFNet contains three convolutional layers and two linear layers for NMNIST classification. The HST attains 84.28\% accuracy on ImageNet, surpassing other current spiking transformer architectures.

The timestep of models for static datasets is set to $4$, and for dynamic datasets is set to $10$. We adopt $\tau=0.5$ and $V_{th}=1$ for all LIF neurons.

{\bf{Training details. }} All experiments are conducted on NVIDIA Tesla A100 GPU with 40GB memory. We train all SNN models with STBP \cite{wu2018spatio} for 600 epochs (static datasets) or 200 epochs (dynamic datasets). We adopt the stochastic gradient descent optimizer with 0.1 learning rate and 0.9 momentum for spiking ResNet-18, and adopt the adam \cite{kingma2014adam} optimizer with 0.001 learning rate for other models. The weight decay is set to 0, and we use the cosine annealing scheduler \cite{loshchilov2017sgdr} to adjust the learning rate. Additionally, TET \cite{deng2022temporal} loss is utilized to improve the accuracy. The seed is set to 0 across all experiments.
\section{Details of Fixed Surrogate Gradients}

In Sec 4.3 of the main text, we conduct extensive experiments to validate the effectiveness of various attack-phase SG, including fixed SGs and our PDSG. The fixed SGs consist of rectangular SG \cite{wu2018spatio}, triangle SG \cite{deng2022temporal}, and ATan SG \cite{fang2021deep}. The rectangular SG is described as:
\begin{equation}
    \frac{\partial s}{\partial u} =
    \begin{cases}
        \frac{1}{2w}, &-w<|u-V_{th}|<w \\
        0, & \text{otherwise}
    \end{cases}.
    \label{eq:suppl_rectangular_sg}
\end{equation}

Here $w$ represents the width of the SG. Typically, $w$ is a hyper-parameter, and we adopt $w=1,2,0.5$ in experiments. The triangle SG is:
\begin{equation}
    \frac{\partial s}{\partial u} = \frac{1}{\gamma^2}\text{max}\{0, \gamma-|u-V_{th}|\}.
    \label{eq:suppl_triangle_sg}
\end{equation}

Here $\gamma$ controls the shape of the SG, and we set $\gamma=1$ in our experiments, which is the default setting in \cite{deng2022temporal}. The ATan SG is denoted as:
\begin{equation}
    \frac{\partial s}{\partial u} = \frac{\alpha}{2(1+(\frac{\pi}{2}\alpha(u-V_{th}))^2)}.
    \label{eq:suppl_atan_sg}
\end{equation}

\begin{figure}[tb]
  \centering
  \includegraphics[height=6cm]{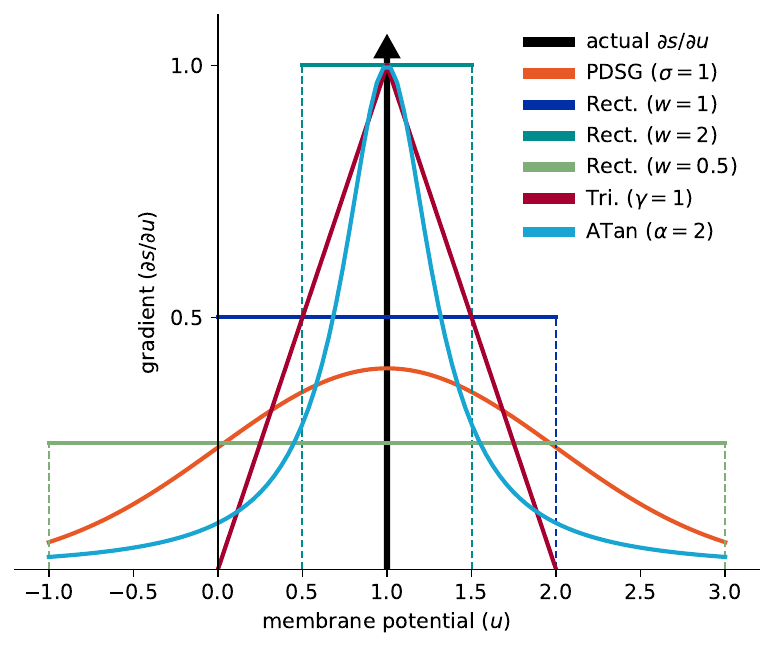}
  \caption{Illustration of various fixed SGs and our PDSG.}
  \label{fig:suppl_fixed_sg}
\end{figure}

We use the default $\alpha=2$ in our experiments. We depict all SGs above and our PDSG in \cref{fig:suppl_fixed_sg}.

\section{Visualization}

\begin{figure*}[tb]
  \centering
  \includegraphics[height=7.8cm]{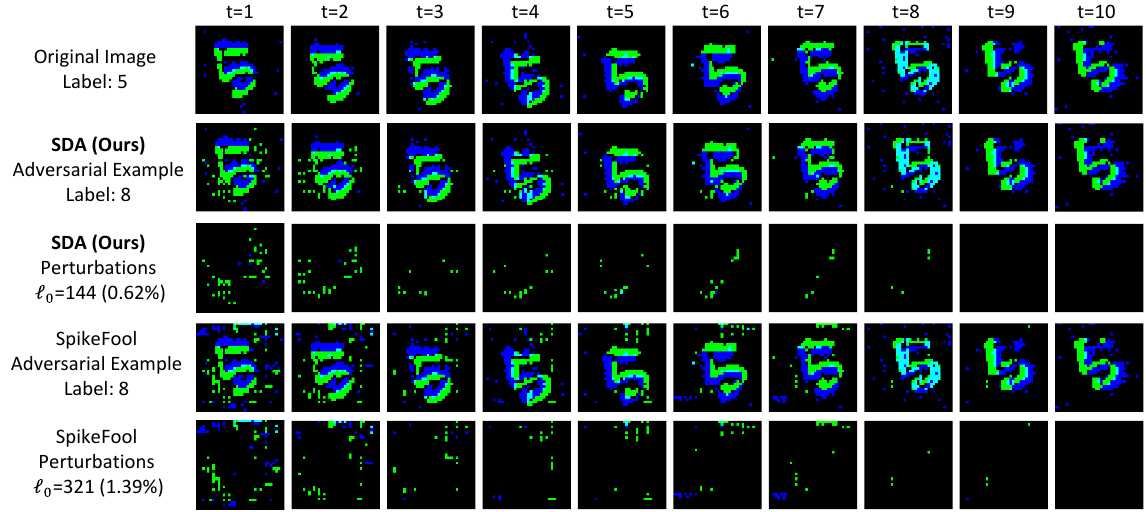}
  \caption{Visualization of the our SDA and SpikeFool on NMNIST dataset. The channel of $p=on$ and $p=off$ is indicated in green and blue color, respectively. Our SDA modifies only 0.62\% of pixels to change the classification result from 5 to 8.}
  \label{fig:suppl_nmnist_visualization}
\end{figure*}

In this section, we present the visualization result of our SDA and the SpikeFool \cite{buchel2022adversarial} in attacking binary dynamic images. The visualization on NMNIST dataset is shown in \cref{fig:suppl_nmnist_visualization}. After attacking, the label of the original image is changed from 5 to 8. Our SDA modifies 144 pixels, which is only 0.62\% of the pixels of the image. In contrast, the SpikeFool modifies 321 pixels, indicating that the perturbations are easier to be detected.

\begin{figure*}[tb]
  \centering
  \includegraphics[height=7.8cm]{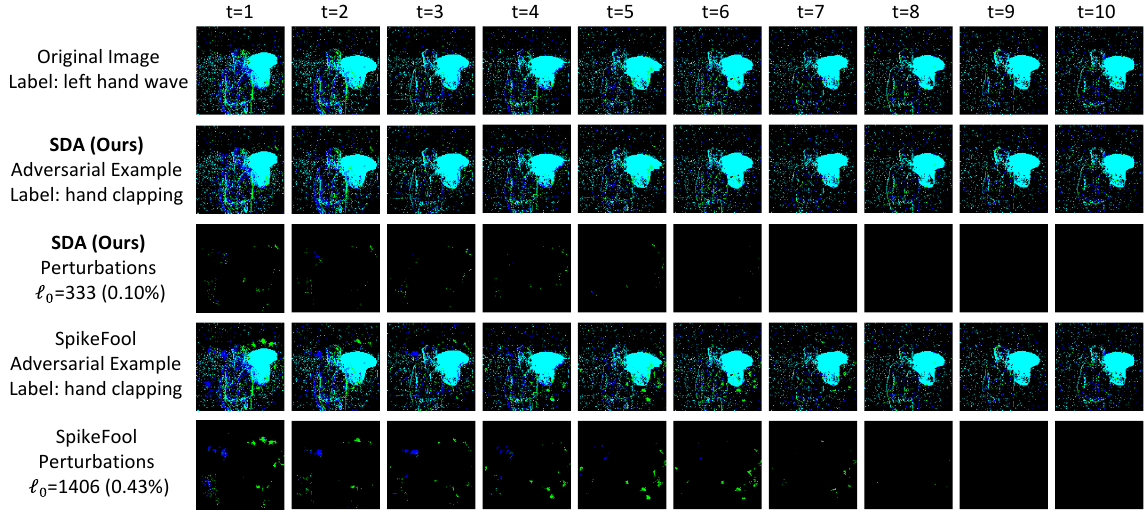}
  \caption{Visualization of the our SDA and SpikeFool on DVS-Gesture dataset. The channel of $p=on$ and $p=off$ is indicated in green and blue color, respectively. Our SDA modifies only 0.10\% of pixels to change the classification result from left hand wave to hand clapping.}
  \label{fig:suppl_dvsgesture_visualization}
\end{figure*}

The visualization on DVS-Gesture dataset is displayed in \cref{fig:suppl_dvsgesture_visualization}. Our SDA only modifies 0.1\% of the pixels, rendering the adversarial example virtually indistinguishable from the original image to both human observers and automated detection systems.

\begin{figure*}[tb]
  \centering
  \includegraphics[height=7.8cm]{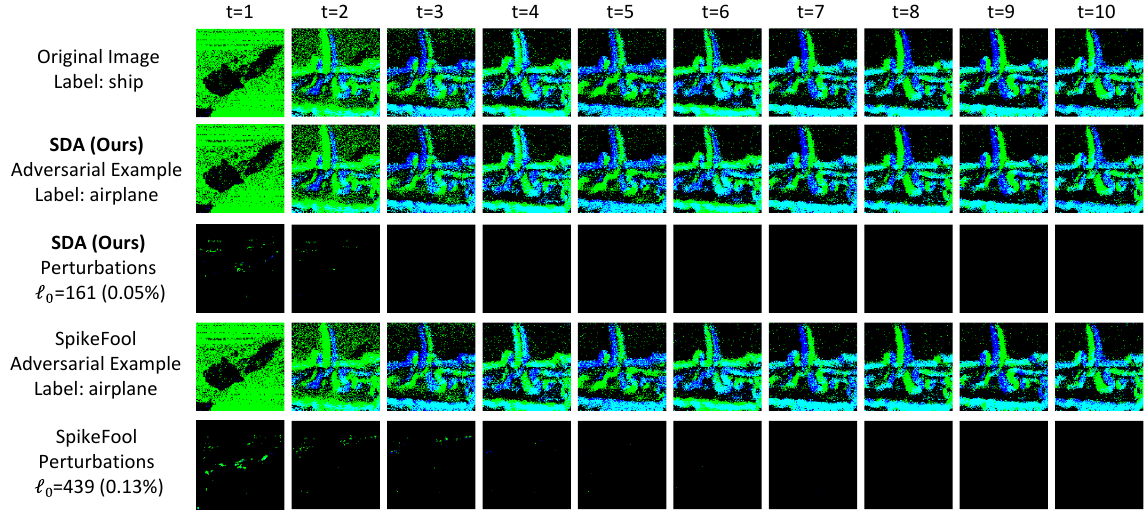}
  \caption{Visualization of the our SDA and SpikeFool on CIFAR10-DVS dataset. The channel of $p=on$ and $p=off$ is indicated in green and blue color, respectively. Our SDA modifies only 0.05\% of pixels to change the classification result from ship to airplane.}
  \label{fig:suppl_cifar10dvs_visualization}
\end{figure*}

We also depict the visualization result on CIFAR10-DVS dataset in \cref{fig:suppl_cifar10dvs_visualization}. In this case, our SDA modifies a mere 0.05\% of the pixels, and the perturbations only exist in the first two frames. Therefore, it suggests that the model focuses on the first two frames to perform classification, and our SDA exploits this behavior to generate imperceptible perturbations.

\section{Discussion of Timesteps in Binary Attack}
\begin{table}[tb]
  \centering
  \setlength\tabcolsep{4.0pt}
  \footnotesize
  \begin{tabular}{@{}ccccccc@{}}
    \hline
    \multirow{3}{*}{T} & \multirow{3}{*}{\makecell{Acc.\\(\%)}} & \multirow{3}{*}{Attack} & Static Evaluation & \multicolumn{3}{c}{Dynamic Evaluation} \\
    \cline{4-7}
    & & & \multirow{2}{*}{\makecell{ASR. (\%)\\($\ell_0<200/800$)}} & \multirow{2}{*}{\makecell{ASR.\\(\%)}} & \multirow{2}{*}{\makecell{Mean\\$\ell_0$}} & \multirow{2}{*}{\makecell{Median\\$\ell_0$}}\\
    & & & & & & \\
    \hline
    \multirow{2}{*}{5} & \multirow{2}{*}{76.5} & SpikeFool & 45.0/99.0 & \textbf{100.0} & 270.24 & 230.00\\
    \cline{3-7}
    & & \textbf{SDA(Ours)} & \textbf{77.0}/\textbf{100.0} & \textbf{100.0} & \textbf{131.08} & \textbf{86.50}\\
    \hline
    \multirow{2}{*}{10} & \multirow{2}{*}{78.2} & SpikeFool & 19.0/70.0 & \textbf{100.0} & 674.89 & 491.00\\
    \cline{3-7}
    & & \textbf{SDA(Ours)} & \textbf{38.0}/\textbf{82.0} & \textbf{100.0} & \textbf{458.02} & \textbf{303.00} \\
    \hline
    \multirow{2}{*}{20} & \multirow{2}{*}{82.4} & SpikeFool & 4.0/11.0 & 72.0 & \textbf{3733.49} & 2705.00\\
    \cline{3-7}
    & & \textbf{SDA(Ours)} & \textbf{5.0}/\textbf{13.0} & \textbf{89.0} & 4905.00 & \textbf{2570.00} \\
  \hline
  \end{tabular}
  \caption{Discussion of attacking spiking ResNet-18  with various timesteps on CIFAR10-DVS dataset. T denotes the timestep, and ASR. denotes the attack success rate. The best results are in bold.}
  \label{tab:suppl_timestep_dynamic}
\end{table}

Since the performance of the SNN model depends on the timestep, we discuss the impact of the timestep in attacking binary dynamic images. As the imperceptibility of the SCG \cite{liang2021exploring} and the GSAttack \cite{yao2024exploring} is insufficient, we only compare our SDA with the SpikeFool \cite{buchel2022adversarial}. The results of attacking spiking ResNet-18 on CIFAR10-DVS dataset is illustrated in \cref{tab:suppl_timestep_dynamic}. Our SDA outperforms the SpikeFool in terms of the attack success rate and sparsity. In timestep $=20$, the difficulty of the attacks increases as the number of the input pixels is large, while our SDA still exhibits stable performance of 89\% ASR. Since we only record $\ell_0$ of successful attack, the mean of $\ell_0$ of our SDA is higher than that of SpikeFool. Notably in timestep $=5$, our SDA achieves a median of 86.5 $\ell_0$, which is only 0.05\% of the input pixels.

\section{Discussion of Initial Selection Count in SDA}

In this section, we conduct experiments with various choices of $k_{init}$ in our SDA. The results are shown in \cref{tab:suppl_k}. We first set $k=k_{init}$ for each iteration, indicating that the $k$ is fixed. As the calculation of the gradients is a course estimation , significant gradients are easy to be ignored when $k$ is fixed at 10, causing low attack success rate. The mean and median of $\ell_0$ is extremely low since we only record $\ell_0$ and count of iterations for successful attacks. Therefore, selecting a fixed low $k$ induces poor attack performance. Conversely, setting a fixed $k=100$ achieves 100\% attack success rate but at the cost of a relatively larger $\ell_0$.

To achieve a stable attack and avoid the hyper-parameter significantly influencing the performance of the attack, we adopt the incremental $k$ strategy in our SDA. The motivation comes from preventing gradient vanishing. In the early stages of the generation process, the model's output is distant from the classification boundary, causing substantial gradients becoming zero. In this case, only a few gradients are valid and we only require to leverage these gradients to calculate their FDs. However, in the later stages, any modified pixel could potentially make the input adversarial, necessitating consideration of a wider range of pixels with contributing gradients. Consequently, we adopt the incremental $k$ in our SDA, indicating that $k$ is incremental by $k_{init}$ in each iteration.
\begin{table}[tb]
  \centering
  \setlength\tabcolsep{3.0pt}
  \footnotesize
  \begin{tabular}{@{}cccccc@{}}
    \hline
    \multirow{3}{*}{$k_{init}$} & Static Evaluation & \multicolumn{4}{c}{Dynamic Evaluation} \\
    \cline{2-6}
    & \multirow{2}{*}{\makecell{ASR. (\%)\\($\ell_0<200/800$)}} & \multirow{2}{*}{\makecell{ASR.\\(\%)}} & \multirow{2}{*}{\makecell{Mean\\$\ell_0$}} & \multirow{2}{*}{\makecell{Median\\$\ell_0$}} & \multirow{2}{*}{\makecell{Mean\\Iterations}}\\
    & & & & &\\
    \hline
    10 (Fixed) & 12.0/12.0 & 12.0 & 13.67 & 12.50 & 3.17\\
    \hline
    20 (Fixed) & 27.0/32.0 & 32.0 & 85.47 & 54.00 & 10.53 \\
    \hline
    50 (Fixed) & 38.0/82.0 & 92.0 & 361.86 & 256.50 & 16.78 \\
    \hline
    100 (Fixed) & 34.0/83.0 & 100.0 & 464.53 & 309.50 & 9.97 \\
    \hline
    1 (Incremental) & 41.0/84.0 & 99.0 & 426.97 & 280.00 & 56.11\\
    \hline
    5 (Incremental) & 38.0/84.0 & 100.0 & 439.55 & 285.50 & 17.82 \\
    \hline
    \textbf{10 (Incremental)} & 38.0/82.0 & 100.0 & 458.02 & 303.00 & 12.33\\
    \hline
    20 (Incremental) & 37.0/80.0 & 100.0 & 466.72 & 314.50 & 8.70 \\
    \hline
    50 (Incremental) & 34.0/75.0 & 100.0 & 518.95 & 341.00 & 5.64 \\
  \hline
  \end{tabular}
  \caption{Attack success rate and dynamic evaluation for attacking spiking ResNet-18 on CIFAR10-DVS dataset with various choices of $k_{init}$ in our SDA. Fixed represents $k$ is equal to $k_{init}$ in each iteration. Incremental denotes $k$ is incremental by $k_{init}$ in each iteration.}
  \label{tab:suppl_k}
\end{table}

As shown in \cref{tab:suppl_k}, the sparsity of perturbations decreases with an increase of $k_{init}$. Since the contributing FDs and reduction process effectively remove redundant perturbations, the $\ell_0$ and attack success rate will not change drastically with variations in $k_{init}$. However, an extremely low $k_{init}$ may cause failed attacks (99\% ASR in $k_{init}=1$). Additionally, a low $k_{init}$ implies that the generation process requires more iterations to find an adversarial example, thus increasing the attack time. Therefore, to make a trade-off between the imperceptibility of perturbations and the time costs of the attack, while ensuring 100\% attack success rate, we choose $k_{init}=10$ in our SDA. 
\begin{table}[tb]
  \centering
  \setlength\tabcolsep{2.5pt}
  \footnotesize
  \begin{tabular}{@{}ccccccc@{}}
    \hline
    \multirow{3}{*}{$\tau$} & \multirow{3}{*}{\makecell{Acc.\\(\%)}} & \multirow{3}{*}{Attack} & \multicolumn{4}{c}{ASR. ($\epsilon=2/255$) / ($\epsilon=8/255$)} \\
    \cline{4-7}
    & & & \multirow{2}{*}{STBP} & \multirow{2}{*}{RGA} & \multirow{2}{*}{HART} & \multirow{2}{*}{\makecell{\textbf{PDSG}\\ \textbf{(Ours)}}}\\
    & & & & & & \\
    \hline
    \multirow{2}{*}{0.25} & \multirow{2}{*}{94.52} & FGSM & 41.81/63.79 & 29.88/49.35 & 42.90/57.80 & \textbf{45.04}/\textbf{82.88}\\
    \cline{3-7}
    & & PGD & 73.48/99.89 & 62.00/96.05 & \textbf{79.68}/99.41 & 70.41/\textbf{99.99}\\
    \hline
    \multirow{2}{*}{0.5} & \multirow{2}{*}{94.72} & FGSM & 38.21/52.36 & 31.14/45.80 & 37.30/46.72 & \textbf{43.98}/\textbf{79.56}\\
    \cline{3-7}
    & & PGD & 66.74/99.81 & 61.97/92.47 & 66.77/98.64 & \textbf{69.62}/\textbf{100.0} \\
    \hline
    \multirow{2}{*}{0.75} & \multirow{2}{*}{94.33} & FGSM & 32.67/45.22 & 26.99/34.47 & 33.62/43.74 & \textbf{42.82}/\textbf{79.64}\\
    \cline{3-7}
    & & PGD & 60.60/99.42 & 48.69/89.30 & 61.16/97.71 & \textbf{68.00}/\textbf{99.96} \\
    \hline    
    \multirow{2}{*}{1.0} & \multirow{2}{*}{94.24} & FGSM & 29.57/40.57 & 27.48/35.42 & 31.52/41.99 & \textbf{43.37}/\textbf{77.98}\\
    \cline{3-7}
    & & PGD & 54.35/95.39 & 48.10/87.99 & 57.23/94.89 & \textbf{68.21}/\textbf{99.94} \\
  \hline
  \end{tabular}
  \caption{Attack success rate for attacking spiking ResNet-18 with various leakage factors on CIFAR10 dataset. $\tau$ denotes the leakage factor. The best results are in bold.}
  \label{tab:suppl_leak_static}
\end{table}

\section{Discussion of Leakage Factors}

To verify the generalization abilities of our PDSG and SDA, we conduct experiments on models with various leakage factors. First, we validate the performance of our PDSG in attacking spiking ResNet-18 on CIFAR10 dataset. The results are illustrated in \cref{tab:suppl_leak_static}. Although our PDSG is surpassed by HART \cite{hao2024threaten} in PGD ($\epsilon=2/255$) attack when $\tau=0.25$, likely due to the compatibility of HART's surrogate function with the model, our PDSG exhibits superior performance in all other experiments.

In \cref{tab:suppl_leak_dynamic}, we demonstrate the performance of our SDA in attacking spiking ResNet-18 on CIFAR10DVS dataset. Our SDA outperforms the SpikeFool across diverse leakage factors. It is noteworthy that the $\ell_0$ of the perturbations increases as the leakage factor decreases, indicating that the model may exhibit greater robustness with a lower leakage factor.
\begin{table}[tb]
  \centering
  \setlength\tabcolsep{3.9pt}
  \footnotesize
  \begin{tabular}{@{}ccccccc@{}}
    \hline
    \multirow{3}{*}{$\tau$} & \multirow{3}{*}{\makecell{Acc.\\(\%)}} & \multirow{3}{*}{Attack} & Static Evaluation & \multicolumn{3}{c}{Dynamic Evaluation} \\
    \cline{4-7}
    & & & \multirow{2}{*}{\makecell{ASR. (\%)\\($\ell_0<200/800$)}} & \multirow{2}{*}{\makecell{ASR.\\(\%)}} & \multirow{2}{*}{\makecell{Mean\\$\ell_0$}} & \multirow{2}{*}{\makecell{Median\\$\ell_0$}}\\
    & & & & & & \\
    \hline
    \multirow{2}{*}{0.25} & \multirow{2}{*}{82.5} & SpikeFool & 18.0/48.0 & \textbf{100.0} & 1263.20 & 896.50\\
    \cline{3-7}
    & & \textbf{SDA(Ours)} & \textbf{27.0}/\textbf{67.0} & \textbf{100.0} & \textbf{639.13} & \textbf{402.00}\\
    \hline
    \multirow{2}{*}{0.5} & \multirow{2}{*}{78.2} & SpikeFool & 19.0/70.0 & \textbf{100.0} & 674.89 & 491.00\\
    \cline{3-7}
    & & \textbf{SDA(Ours)} & \textbf{38.0}/\textbf{82.0} & \textbf{100.0} & \textbf{458.02} & \textbf{303.00} \\
    \hline
    \multirow{2}{*}{0.75} & \multirow{2}{*}{78.0} & SpikeFool & 38.0/87.0 & \textbf{100.0} & 374.43 & 271.00\\
    \cline{3-7}
    & & \textbf{SDA(Ours)} & \textbf{57.0}/\textbf{92.0} & \textbf{100.0} & \textbf{261.16} & \textbf{152.50} \\
    \hline
    \multirow{2}{*}{1.0} & \multirow{2}{*}{76.9} & SpikeFool & 39.0/97.0 & \textbf{100.0} & 309.43 & 253.00\\
    \cline{3-7}
    & & \textbf{SDA(Ours)} & \textbf{67.0}/\textbf{99.0} & \textbf{100.0} & \textbf{175.59} & \textbf{105.50} \\
  \hline
  \end{tabular}
  \caption{Attack success rate and dynamic evaluation for models with various leakage factors on binary dynamic images. $\tau$ denotes the leakage factor. The best results are in bold.}
  \label{tab:suppl_leak_dynamic}
\end{table}

\section{Comparison with Black-box Attacks}

In contrast to white-box attacks, black-box attacks also threaten neural network models. Without accessing the weights and architectures of the models, black-box attacks only require the inputs and outputs of models, and leverage them to generate adversarial examples. Transfer-based black-box attacks are already evaluated in \cite{bu2023rate, hao2024threaten}. To validate our PDSG's ability of optimizing the gradient flow, we conduct experiments with score-based black-box Square Attack \cite{andriushchenko2020square} and decision-based black-box attack RayS \cite{chen2020rays} in \cref{tab:rebut_blackbox}. The results demonstrate that our PDSG outperforms black-box attacks across various models and datasets, except adversarially trained models. As ResNet18 is specifically adversarially trained by PGD attack with $\epsilon=2/255$, the PDSG performs poorly when $\epsilon=2/255$. However, when the attack intensity increases to $\epsilon=8/255$, our PDSG surpasses other black-box attacks.

\begin{table}[tb]
  \centering
  \setlength\tabcolsep{1.3pt}
  \footnotesize
  \begin{tabular}{@{}cccccccc@{}}
    \hline
    \multirow{3}{*}{\makecell{Dataset}} & \multirow{3}{*}{\makecell{Architecture}} & \multicolumn{3}{c}{ASR. (\%) ($\epsilon=2/255$)} & \multicolumn{3}{c}{ASR. (\%)($\epsilon=8/255$)}\\
    \cline{3-8}
    & & \multirow{2}{*}{\makecell{{\bf PDSG}\\(PGD)}} & \multirow{2}{*}{\makecell{Square}} & \multirow{2}{*}{\makecell{RayS}} & \multirow{2}{*}{\makecell{{\bf PDSG}\\(PGD)}} & \multirow{2}{*}{\makecell{Square}} & \multirow{2}{*}{\makecell{RayS}} \\
    & & & & & & \\
    \hline
    \multirow{4}{*}{\makecell{CIFAR10}} & ResNet18  & {\bf 69.62} & 29.79 & 13.40 & {\bf 100.00} & 66.10 & 52.96 \\
    \cline{2-8}
    & \multirow{2}{*}{\makecell{ResNet18\\(Adv. trained)}} & \multirow{2}{*}{\makecell{10.68}} & \multirow{2}{*}{\makecell{{\bf 44.79}}} & \multirow{2}{*}{\makecell{18.10}} & \multirow{2}{*}{\makecell{{\bf 62.16}}} & \multirow{2}{*}{\makecell{56.54}} & \multirow{2}{*}{\makecell{33.30}} \\
    & & & & & & &\\
    \cline{2-8}
    & VGG11 & {\bf 39.20} & 26.86 & 12.81 & {\bf 99.71} & 57.42 & 56.92 \\
    \hline
    CIFAR100 & ResNet18 & {\bf 78.50} & 50.67 & 32.64 & {\bf 99.83} & 78.80 & 71.80\\
  \hline
  \end{tabular}
  \caption{Attack success rate under comparison with black-box attacks. All inputs adopt direct coding. The best results are in bold.}
  \label{tab:rebut_blackbox}
\end{table}

\section{Results of Adaptive Attack}

In attacking static images, we conduct experiments of APGD \cite{croce2020reliable} attack, which is an adaptive version of the PGD attack. In \cref{tab:rebut_apgd}, the results show the same trend as the PGD attack in \cref{tab:static}, demonstrating that our PDSG performs the best and has stable performance.

\begin{table*}[tb]
  \centering
  \setlength\tabcolsep{5.0pt}
  \footnotesize
  \begin{tabular}{@{}cccccccccc@{}}
    \hline
    \multirow{3}{*}{Dataset} & \multirow{3}{*}{Architecture} & \multicolumn{4}{c}{ASR. (\%) ($\epsilon=2 / 255$)} & \multicolumn{4}{c}{ASR. (\%) ($\epsilon=8 / 255$)}\\
    \cline{3-10}
    & & \multirow{2}{*}{STBP} & \multirow{2}{*}{RGA} & \multirow{2}{*}{HART} & \multirow{2}{*}{\makecell{{\bf{PDSG}}\\{\bf{(Ours)}}}} & \multirow{2}{*}{STBP} & \multirow{2}{*}{RGA} & \multirow{2}{*}{HART} & \multirow{2}{*}{\makecell{{\bf{PDSG}}\\{\bf{(Ours)}}}} \\
    & & & & & & & & & \\
    \hline
    \multirow{4}{*}{\makecell{CIFAR10}} & ResNet18 & 71.36 & 67.04 & 71.49 & {\bf 75.18} & 99.67 & 94.77 & 98.54 & {\bf 99.97} \\
    \cline{2-10}
    & \multirow{2}{*}{\makecell{ResNet18\\(Adv. trained)}} & \multirow{2}{*}{\makecell{14.34}} & \multirow{2}{*}{\makecell{17.93}} & \multirow{2}{*}{\makecell{21.20}} & \multirow{2}{*}{\makecell{{\bf 21.56}}} & \multirow{2}{*}{\makecell{41.39}} & \multirow{2}{*}{\makecell{57.37}} & \multirow{2}{*}{\makecell{70.74}} & \multirow{2}{*}{\makecell{{\bf 71.92}}} \\
    & & & & & & & & & \\
    \cline{2-10}
    & VGG11 & 46.96 & 46.40 & {\bf 54.82} & 45.58 & 99.25 & 88.51 & 98.13 & {\bf 99.84} \\
    \hline
    CIFAR100 & ResNet18 & 85.12 & 82.62 & {\bf 89.54} & 83.21 & 99.68 & 98.62 & 99.67 & {\bf 99.91} \\
  \hline
  \end{tabular}
  \caption{Comparison with state-of-the-art approaches on attacking static images using APGD attack. ASR. denotes the attack success rate. $\epsilon$ is the attack intensity. STBP denotes attacking using training-phase SG. All inputs adopt direct coding. The best results are in bold.}
  \label{tab:rebut_apgd}
\end{table*}

\section{Evaluation of Computational Cost}

To evaluate the computational cost of our method, we adopt $batch\_size=1$ and perform attacks on both static and dynamic datasets. In \cref{tab:rebut_time_cost_static}, since our PDSG requires the computation of the standard deviation of membrane potential, the efficiency of our PDSG is slightly lagging behind. As shown in \cref{tab:rebut_time_cost_dynamic}, when attacking binary dynamic images, our SDA performs more efficiently than SpikeFool. Although SCG and GSAttack execute fast, their $\ell_0$ are much larger than ours. Specifically, when cooperating with the PDSG, our SDA achieves a significant efficiency improvement, since the PDSG optimizes the gradient flow and effectively reduces the number of iterations.
\begin{table}[tb]
  \centering
  \setlength\tabcolsep{5.0pt}
  \footnotesize
  \begin{tabular}{@{}ccccccc@{}}
    \hline
    \multirow{3}{*}{\makecell{Dataset}} & \multirow{3}{*}{\makecell{Architecture}} & \multirow{3}{*}{\makecell{Attack}} & \multicolumn{4}{c}{Attack time per sample (s)} \\
    \cline{4-7}
    & & & \multirow{2}{*}{\makecell{STBP}} & \multirow{2}{*}{\makecell{RGA}} & \multirow{2}{*}{\makecell{HART}} & \multirow{2}{*}{\makecell{{\bf PDSG}\\{\bf (Ours)}}} \\
    & & & & & & \\
    \hline
    \multirow{2}{*}{\makecell{CIFAR10}} & \multirow{2}{*}{\makecell{ResNet18}} & FGSM & 0.33 & 0.31 & 0.40 & 0.55 \\
    \cline{3-7}
    & & PGD & 2.15 & 1.58 & 2.08 & 3.22 \\
  \hline
  \end{tabular}
  \caption{Computational costs in attacking static images.}
  \label{tab:rebut_time_cost_static}
\end{table}
\begin{table}[tb]
  \centering
  \setlength\tabcolsep{2.5pt}
  \footnotesize
  \begin{tabular}{@{}ccccccc@{}}
    \hline
    \multirow{3}{*}{\makecell{Dataset}} & \multirow{3}{*}{\makecell{Architecture}} & \multirow{3}{*}{\makecell{Gradient}} & \multicolumn{4}{c}{Attack time per sample (s)} \\
    \cline{4-7}
    & & & \multirow{2}{*}{\makecell{SCG}} & \multirow{2}{*}{\makecell{SpikeFool}} & \multirow{2}{*}{\makecell{GSAttack}} & \multirow{2}{*}{\makecell{{\bf SDA}\\{\bf (Ours)}}} \\
    & & & & & & \\
    \hline
    \multirow{3}{*}{\makecell{N-MNIST}} & \multirow{3}{*}{\makecell{PLIFNet}} & STBP & 0.26 & 12.44 & 3.65 & 7.48 \\
    \cline{3-7}
    & & \multirow{2}{*}{\makecell{{\bf PDSG}\\{\bf (Ours)}}} & \multirow{2}{*}{\makecell{0.24}} & \multirow{2}{*}{\makecell{27.18}} & \multirow{2}{*}{\makecell{2.14}} & \multirow{2}{*}{\makecell{0.95}} \\
    & & & & & & \\
  \hline
  \end{tabular}
  \caption{Computational costs in attacking binary dynamic images.}
  \label{tab:rebut_time_cost_dynamic}
\end{table}
\end{document}